\documentclass{article}

\usepackage[sort&compress,square,numbers]{natbib}
\usepackage[final]{corl_2020} %
\usepackage[font=footnotesize]{caption}
\usepackage{natbib}
\usepackage[english]{babel}
\usepackage{blindtext}
\usepackage{todonotes}
\usepackage{times}
\usepackage{epsfig}
\usepackage{graphicx}
\graphicspath{{./images/}}
\usepackage{amsmath}
\DeclareMathOperator*{\argmax}{argmax}
\usepackage{amssymb}
\usepackage{array}
\usepackage{tabulary}
\usepackage{dblfloatfix}
\usepackage{booktabs}
\usepackage{tikz}
\usetikzlibrary{positioning}
\usepackage{flushend}
\usepackage{caption}
\usepackage{subcaption}
\usepackage{microtype}
\usepackage{wrapfig}
\usepackage{multicol}
\usepackage[subtle]{savetrees}

\title{\LARGE \bf
PLOP: Probabilistic poLynomial Objects trajectory Prediction for autonomous driving
}

\author{Thibault Buhet$^{1}$, Emilie Wirbel$^{1, 2}$, Andrei Bursuc$^{2}$ and Xavier Perrotton$^{1}$ \\
$^{1}$Valeo Driving Assistance Research, $^{2}$ Valeo.ai \\
\texttt{name.surname@valeo.com}
}

\begin{document}

\maketitle
\thispagestyle{empty}
\pagestyle{empty}

\begin{abstract}

To navigate safely in urban environments, an autonomous vehicle (\emph{ego vehicle}) must understand and anticipate its surroundings, in particular the behavior and intents of other road users (\emph{neighbors}). Most of the times, multiple decision choices are acceptable for all road users (e.g., turn right or left, or different ways of avoiding an obstacle), leading to a highly uncertain and multi-modal decision space. We focus here on predicting multiple feasible future trajectories for both ego vehicle and neighbors through a probabilistic framework. We rely on a conditional imitation learning algorithm, conditioned by a navigation command for the ego vehicle (e.g., ``turn right''). Our model processes ego vehicle front-facing camera images and bird-eye view grid, computed from Lidar point clouds, with detections of past and present objects, in order to generate multiple trajectories for both ego vehicle and its neighbors. Our approach is computationally efficient and relies only on on-board sensors. We evaluate our method offline on the publicly available dataset nuScenes, achieving state-of-the-art performance, investigate the impact of our architecture choices on online simulated experiments and show preliminary insights for real vehicle control.

\end{abstract}

\section{Introduction}

\begin{wrapfigure}[23]{r}{0.5\textwidth}
\renewcommand{\captionfont}{\footnotesize}
\centering
\includegraphics[width=0.5\textwidth]{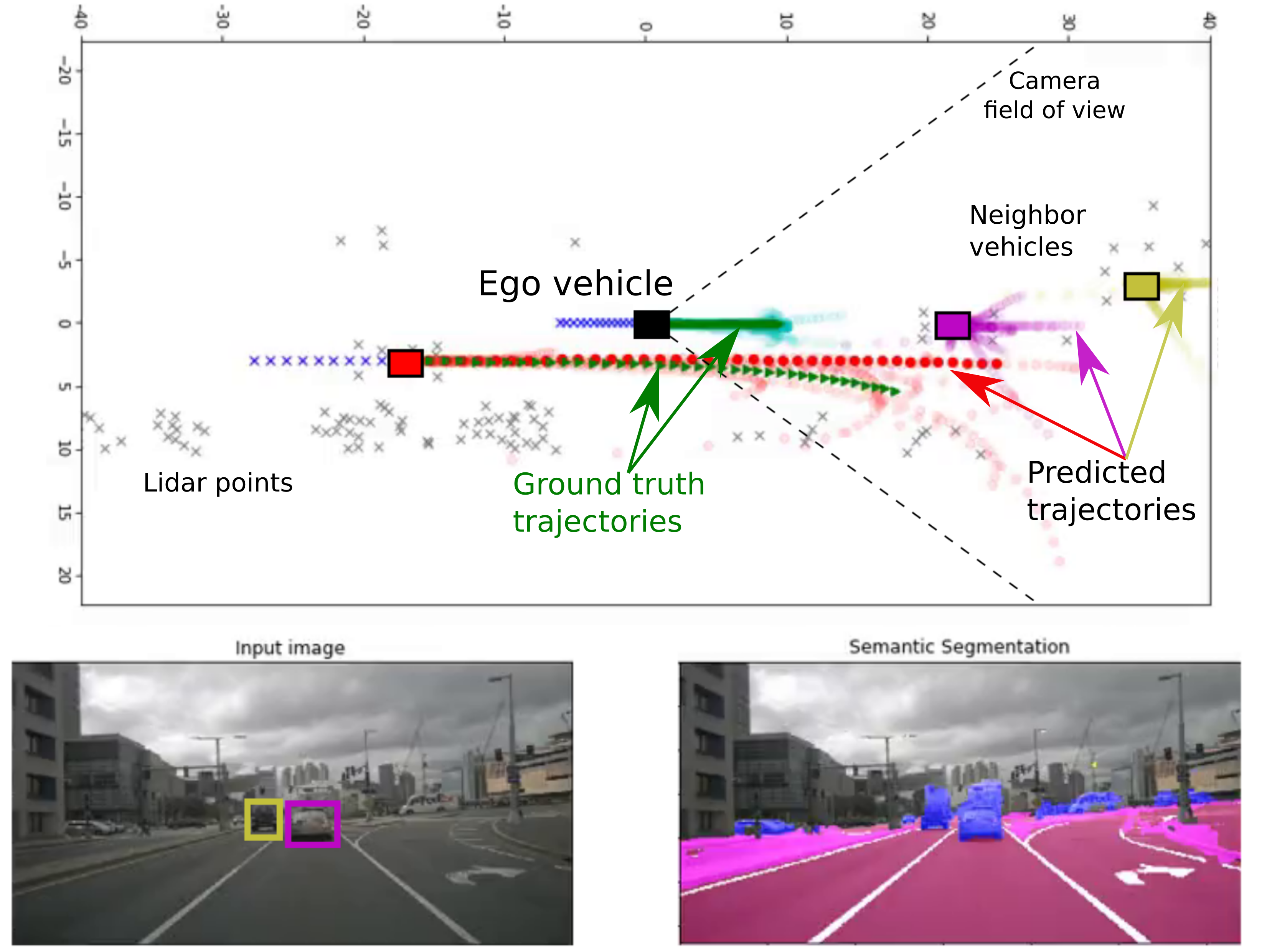}
 
 \caption{\textbf{Qualitative example of trajectory predictions on a test sample from nuScenes dataset.} PLOP processes front camera image, Lidar points, 2s of ego vehicle past track and 1s of neighbor vehicles past tracks to perform trajectory prediction over the next 4s. PLOP handles uncertainty and variability by predicting vehicle trajectories as a probabilistic Gaussian Mixture, constrained by a polynomial formulation.}
 \label{fig:yield}
\end{wrapfigure}

Operating self-driving cars in the real world is a highly challenging endeavor. The vehicle must interact safely with other road users and stick to the limits and rules of the road. Most human drivers do this effortlessly through visual perception and driving experience allowing them to quickly scan surroundings and perform various micro-decisions needed in traffic. For a vehicle, we consider that over a short enough time interval, the world can be approximated through a snapshot of the current scene in which agents will take actions.
The static environment might be hard to understand because of the current topology (\textit{e.g.}, complex intersections) or unusual circumstances (\textit{e.g.}, work zone, absent/inconsistent markings, etc.). Other agents might also be tough to handle because they are out of the autonomous vehicle's control. They can be of very different types (\textit{e.g.}, pedestrian, cyclist, car, truck, robot, etc.) and can be involved in 
unusual situations that should be managed (\textit{e.g.}, pedestrian on the highway, animal crossing the road, emergency vehicle intervention, etc.). The autonomous vehicle, which we designate as the \textit{ego vehicle} in the following, will also have a destination to reach and will be guided by either a target position or simply by a high-level goal such as keeping its lane, turning at an intersection etc. We refer to the other vehicles close to the ego vehicles as \textit{neighbors}.

Driving algorithms must follow a correct behavior under these varying and evolving parameters.
\textit{Trajectory prediction} attempts to 
mitigate this problem. 
Here, we define trajectory prediction as the prediction of future positions of all agents in the scene, ego and neighbor vehicles, over a fixed period of time.
Typical real-world driving environments involve high uncertainty and multiple possible outcomes. In such contexts, a trajectory prediction algorithm must consider diverse potential future paths for each agent and estimate a measure of uncertainty on the predictions. We assume the ego vehicle has access to information describing the current scene, \textit{e.g.}, sensor data, object bounding boxes and/or past positions (tracks), semantic
segmentation maps, depth information, etc. 
The input representation of the scene is essential and can vary significantly across related methods: using only raw sensor data~\citep{faf, intentnet}, adding object detections and past positions~\citep{precog, r2p2, convsocialpooling, sociallstm, ConditionalVT, seq2seq}, assuming almost complete explicit information about the scene (semantic, lanes, detections, map, etc.)~\citep{chauffeutnet}. Similar to previous works, we opt for using raw sensor data (front camera and Lidar) and object detections that can be computed on the fly on the vehicle. We forego using HD maps, which albeit useful, bring a significant cost for updating and processing, while still having blind spots, \textit{e.g.}, recent road works.

To this effect, we propose an architecture for multi-modal trajectory prediction, dubbed PLOP (\textbf{P}robabilistic po\textbf{L}ynomial \textbf{O}bjects trajectory \textbf{P}rediction). PLOP predicts multiple plausible trajectories separately for ego vehicle and neighbors, in a single step fashion, through a formalism based on Mixture Density Networks (MDN)~\cite{bishop1994mixture}. \citet{Makansi_2019_CVPR} showed that MDNs can provide stable results with careful training and sampling, in a pedestrian future position prediction context. Here, MDNs coupled with a polynomial formulation for trajectories naturally deal with the ambiguity of agents' behaviours offering multiple realistic futures. PLOP leverages scene information from sensor data (Lidar and front camera) and detection tracks of nearby vehicles, which can be computed from side cameras and/or Lidar point clouds. We show that adding an auxiliary perception task, e.g. semantic segmentation on front image, contributes to further improvement of prediction accuracy through injection of scene layout information in the internal representations of the network. In Figure~\ref{fig:yield} we illustrate typical input data and results for our method. Additional results can be found in the supplementary material video.

The main contributions of this work are the following:
\textbf{(1)} We propose a single-shot, anchor-less trajectory prediction method, based on MDNs and polynomial trajectory constraints, relying only on on-board sensors (no HD map requirement). The polynomial formulation ensures that the predicted trajectories are coherent and smooth, while providing more learning flexibility through the extra parameters. We find that this mitigates training instability and mode collapse that are common to MDNs~\citep{cui2019multimodal};
\textbf{(2)} We extensively evaluate PLOP and show its effectiveness across datasets and settings. We conduct a comparison showing the improvement over state-of-the-art PRECOG~\citep{precog} on the public dataset nuScenes \citep{nuscenes};
\textbf{(3)} We study closed loop performance for the ego vehicle, on simulation and with preliminary insights for real vehicle control.

\section{Related Work}

In contrast to explicit mathematical models \citep{sfmod}, data-driven methods offer the advantage that the models no longer need to be specified explicitly. Recent deep learning methods 
address this problem mostly through Imitation Learning (IL).
We focus here on a subset of IL, behavioral cloning (BC), a supervised learning approach where expert samples are used as ground truth. 
Several datasets such as nuScenes~\citep{nuscenes} or Waymo~ \citep{waymo} %
enable the development of BC algorithms for trajectory prediction. 
We note that for dealing with multiple possible outcomes, there is practically no dataset offering such situations (except for a recent synthetic pedestrian dataset~\citep{liang2020garden}), further increasing the difficulty of this endeavor.
Most of the existing literature assumes that the tracked past positions are known, often by accumulating past object detections. Approaches are usually recursive and attempt to encode the relationships between agents of the road.

There are two main fields of trajectory prediction methods: predicting vehicle behavior, or pedestrians. In most pedestrian prediction problems, the scene is observed from a fixed position with actors entering and leaving. A recent survey by \citet{rudenko2020human} presents of very complete panorama of the state of the art. Since precursor work such as SocialLSTM \cite{sociallstm}, later extended for vehicle applications by CS-LSTM \citep{convsocialpooling}, pedestrian oriented literature often focuses on interactions between agents.
Pedestrian trajectory predictions are often based on graph structures such as the recent Trajectron \cite{ivanovic2019trajectron,salzmann2020trajectron++} or Social-BiGAT \cite{kosaraju2019social}. Here, we will concentrate on vehicle trajectory prediction.

For road users, the scene is usually centered around an ego vehicle. There is a fundamental distinction between predicting the trajectory for the ego vehicle, and predicting the neighbors. In the last case, there is no information available on the goal of the vehicles, and they cannot be controlled. Here, benchmarks such as Argoverse \cite{argoverse} or nuScenes \cite{nuscenes} based on offline datasets can be used, although global metrics do have their limits because they do not distinguish between critical use cases, \textit{e.g.} braking at a red light, and non-critical use cases, \textit{e.g.} lane keeping without obstacle.
In this context, graph representations can also be leveraged for better modeling of interactions between agents for trajectory prediction \citep{stgat, trafficpredict, spectral}. SpAGNN~\cite{casas2019spatially} extend graph representations with detections of different objects in the scene and generate trajectories through iterative message passing instead of a classic RNN-based decoder. In most cases, it is assumed that neighbors are detected tracked, but some approaches, such as PnPNet \cite{liang2020pnpnet} learn jointly perception, object track association, and future trajectory prediction from HD maps and Lidar frames.

Our work is focused on predicting the ego vehicle trajectory, with a closed loop control application in mind, along with the neighbors trajectories. Some approaches such as ChauffeurNet~\citep{chauffeutnet} use a high-level scene representation (road map, traffic lights, speed limit, route, dynamic bounding boxes, etc.). More recently, MultiPath~\citep{multipath} uses trajectory anchors, used in one-step object detection, extracted from the training data for ego vehicle prediction. \citet{rules} use a high level representation which includes some dynamic context. \citet{cui2019multimodal} produce multi-modal Gaussian representations, however contrary to our work the modes are constrained by the possible manoeuvres, e.g. turn at an intersection.

In contrast, we choose to leverage also low level sensor data, here Lidar point clouds and camera image. In that domain, recent approaches address the variation in agent behaviors by predicting multiple trajectories, often in a stochastic way.
Many works, \textit{e.g.}, PRECOG~\citep{precog}, R2P2~\citep{r2p2}, 
MFP~\citep{mfp}, SocialGAN~\citep{socialgan} and others~\citep{deepim}, focus on this aspect 
through a probabilistic framework on the network output or latent representations, producing multiple trajectories for ego vehicle, nearby vehicles or both. \citet{phan2020covernet} generate a trajectory set, then classify correct trajectories. \citet{marchetti2020mantra} generate multiple futures 
from encodings of similar trajectories stored in a memory. \citet{ohn2020learning} learn a weighted mixture of expert policies trained to mimic agents with specific behaviors. 
In PRECOG, \citet{precog} advance a probabilistic formulation that explicitly models interactions between agents, using latent variables to model their plausible reactions, with the possibility to pre-condition the trajectory of the ego vehicle by a goal.

Compared to related works, an advantage of our approach is that we rely on on-board raw sensor data, differently from others relying on high precision maps and GPS positioning. PLOP is trainable end-to-end from imitation learning, where data is relatively easier to obtain. PLOP is computationally efficient during both training and inference as it predicts trajectory coefficients in a single step, without requiring a RNN-based decoder. The polynomial function trajectory coefficients eschew  the need for anchors~\citep{multipath}, whose quality can vary across datasets.

\section{Network Structure}

\begin{wrapfigure}[15]{r}{0.5\textwidth}
    \renewcommand{\captionfont}{\footnotesize}
    \centering
    \vspace{-18pt}
    \includegraphics[width=0.48\textwidth]{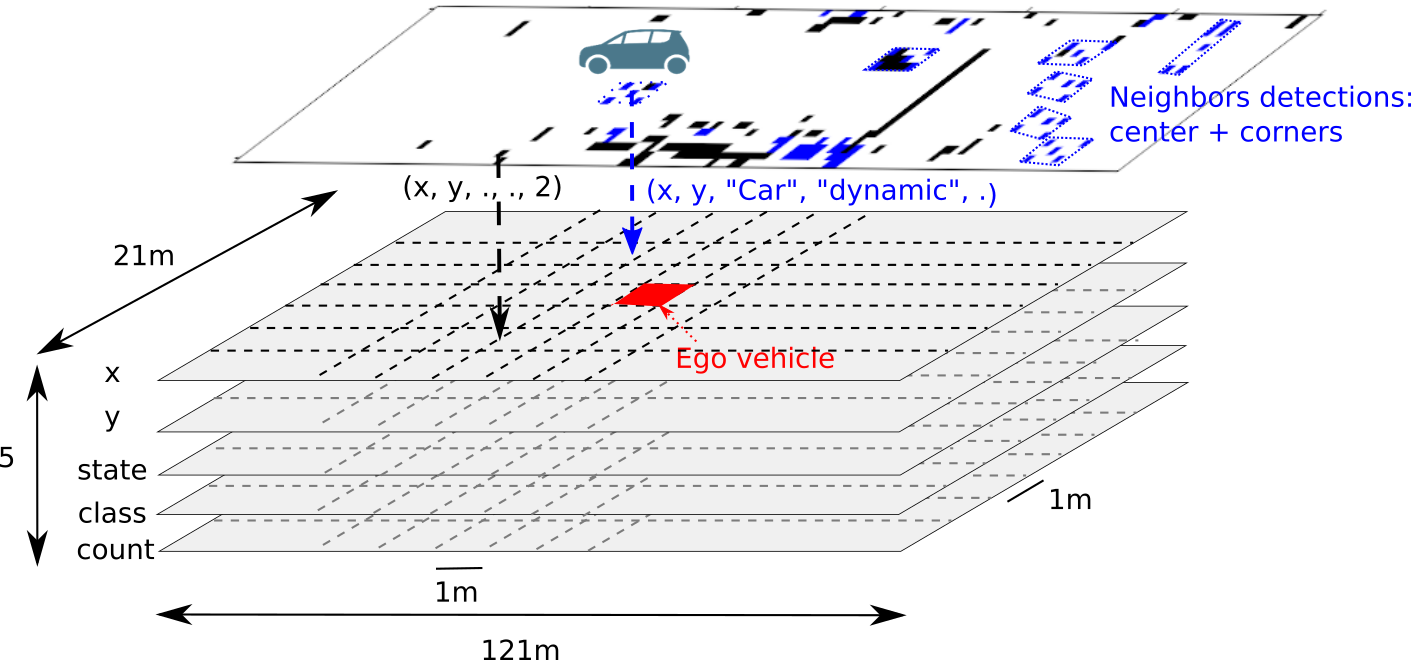}
    \caption{\textbf{Structure and construction of the input birdeye view (BEV):}  vehicles (corners and center) and Lidar points are reprojected in the ego vehicle coordinate system. The BEV ranges from $-60.5$m to $+60.5$m on the longitudinal axis and $-10.5$m to $+10.5$m on the lateral axis using $1$m$\times$ $1$m cells. Each cell contains if applicable the position (resp. mean position) $(x,y)$ of neighbor vehicle (resp. Lidar points), state (parked, stopped, dynamic), class (2wheels, car, truck) and Lidar points count.  Each BEV is represented by a $[121 \times 21 \times 5]$ tensor.}
    \label{fig:input_bev}
\end{wrapfigure}

Our main goal is to predict future ego vehicle trajectories along with the neighbors trajectories. 
To this end, we use a single multi-input multi-output neural network to produce a probabilistic
representation of these trajectories. In this section, we first describe the input structure, then the architecture of the network, and finally the formulation of the outputs and the associated losses.

\subsection{Inputs: Past Trajectories, Camera and Lidar}

We assume all input data is sampled at 10Hz. We note $N$ the maximum number of neighbors that are considered as input ($N=10$ in this work). Past inputs are accumulated over 2s. For simplicity, we only consider vehicles as neighbors, excluding other road users such as pedestrians or bikes from the input object detections.

Past trajectories are represented as time series, over the last 2s %
for the ego vehicle and the neighbors. We use a frontal RGB camera (with a field of view of $70^{\circ}$ on nuScenes).%
The metric surroundings (Lidar point cloud and neighbors detections) are projected into a grid bird-eye view (BEV), inspired by \citep{DOGP, intentnet, precog}, as summarized in Figure~\ref{fig:input_bev}. The ground Lidar points are filtered out for unequivocal obstacle representation. This map is accumulated over the past 2s, \textit{i.e.}, 20 frames.%

Finally, we use a navigation command input for the conditional part of our network \citep{cil}. There are four navigation commands, one when the ego vehicle is far away from an intersection: \textit{follow} and three when the ego vehicle is close to an intersection: \textit{left, straight, right}.

\subsection{Network Architecture}

\begin{wrapfigure}[22]{r}{0.6\textwidth}
 \renewcommand{\captionfont}{\footnotesize}
 \centering
 \includegraphics[width=0.6\textwidth]{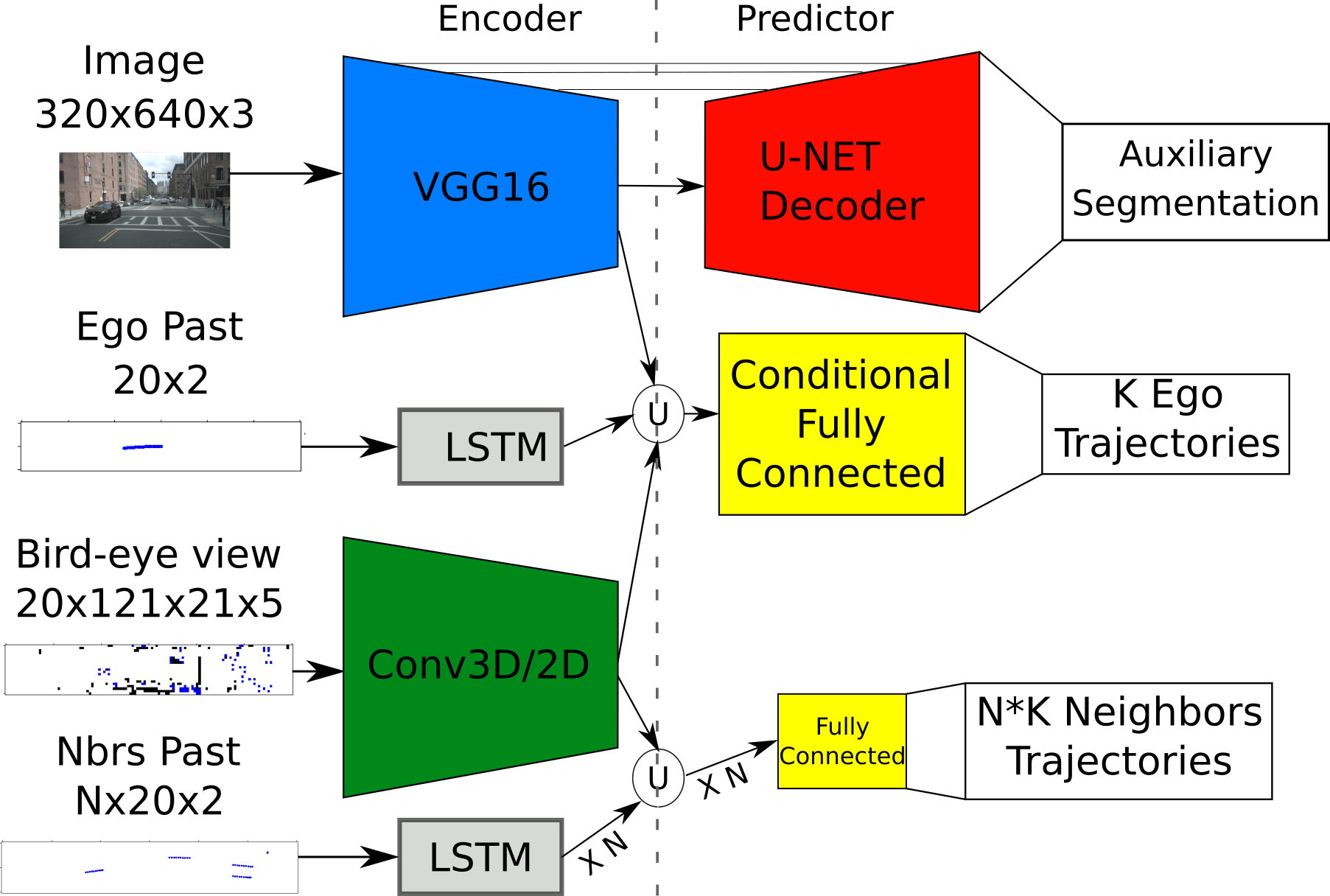}
 \caption{\textbf{Neural network architecture for PLOP:} the encoder is on the left and the predictor is on the right. $\cup$: Concatenation. PLOP is compatible with an arbitrary number of neighbors $N$, though for implementation constraints maximal $N$ is fixed to 10 for training. PLOP architecture is presented in greater details in Section 1 of the supplementary material.}
 \label{fig:network}
\end{wrapfigure}

We illustrate the structure of our neural network in Figure~\ref{fig:network} and detail its design here. The architecture has two main sections: an encoder to synthesize information and a predictor to  process it. Overall, the structure of the neural network is designed to guide latent representations towards encompassing relevant information about the environment and its geometry.

The encoder consists of three parts: the front camera image of the ego vehicle is encoded by a VGG16~\citep{vgg} network, the BEVs are encoded by a CNN with 3D convolutional layers, and the trajectories are encoded using an LSTM layer~\citep{hochreiter1997long}. The LSTM weights are shared between the neighbors, while the ego vehicle has its own LSTM weights. 
The predictor can also be divided in three parts.
First, we consider an auxiliary U-net semantic segmentation decoder~\citep{unet} to inject an awareness of scene layouts and human interpretability in the network features and to improve learning stability of the entire architecture. Note that we are not actually interested in the semantic segmentation prediction itself, but rather use it as pretext task to enable the network to learn useful features. We can remove this module after training.
Then, the neighbors' trajectories prediction takes as input, for each vehicle, a concatenation of the BEV encoding and the considered neighbor vehicle past trajectory encoding and apply 3 fully connected (FC) layers to output a multivariate Gaussian mixture for a fixed number of possible trajectories $K$. 
We train this module with the negative log-likelihood loss~\citep{bishop1994mixture} (see Section \ref{sec:output-and-loss}).
The 
FC weights are shared between all neighbors vehicles. 
Finally the ego vehicle trajectory prediction uses the same principle as for the neighbors vehicles while adding the image encoding as an input and a conditional dimension to the
FC layers. The 3 FC layers are replaced by $4\times 3$ FC layers conditioned by the 4 navigation commands.

We note that each vehicle prediction does not have direct access to the sequence of past positions of other vehicles. The bird's eye view (BEV) encoding implicitly 
encapsulates the interactions between vehicles. It allows our architecture to be agnostic to the number of considered neighbors, 
an advantage compared to methods like Social-LSTM or PRECOG. A bird's eye view is also a lightweight representation for point cloud data, and can be easily complemented with additional information as described in Figure~\ref{fig:input_bev}.

\subsection{Network Outputs and Training}
\label{sec:output-and-loss}

\paragraph{Trajectory prediction}

Our network has two main outputs, the possible future trajectories for both ego vehicle and nearby vehicles. For each vehicle we want to predict multiple trajectories to 
mimic the stochasticity of human behavior
and cope with ambiguities in a given situation. We want to predict a fixed number $K$  of possible trajectories for each vehicle, and associate them to a probability distribution over $x$ and $y$ ($x$ is the longitudinal axis, $y$ the lateral axis, pointing left). For the ego vehicle, we estimate the probability distribution conditioned by the navigation command $c$. For simplification, we consider that $x$ and $y$ are independent. We make the following assumptions and simplifications: \textbf{(1) The distribution is expressed only at fixed points}, sampled at a fixed rate in the future (indexed by $t \in [0, T]$). We forecast trajectories over a 4s horizon, 
\textit{i.e.}, $T=40$ 
here; \textbf{(2)}  For $x$ and $y$ respectively, for each point in time, \textbf{the distribution is modeled by a mixture with $K$ Gaussian components} $\mathcal{N}\big(\hat{\mu}_{k, x}(t), \sigma_{k, x, t} \big)$. By default, we use $K=12$, following the choice of \citet{precog}; \textbf{(3)  The mixture weights $\pi_k$ are shared} for all sampled points belonging to the same trajectory (over $x$ and $y$). %
This makes it possible to associate a weight to a whole trajectory and reduces the number of parameters; \textbf{(4)} For $x$ and $y$ respectively, for each component, the means of the distribution $\hat{\mu}_{k, x}(t)$ (resp. $\hat{\mu}_{k, y}(t)$) are \textbf{generated polynomials of degree 4 of time}, following \citet{ConditionalVT}.  We denote the coefficients of these polynomials $\boldsymbol{a}_{k,d,x}$ (resp. $\boldsymbol{a}_{k,d, y}$), the constant coefficient is set to zero. This  reduces the number of parameters and constrains the dynamics of the points produced by the trajectories.%

\begin{wrapfigure}[14]{r}{0.4\textwidth}
    \renewcommand{\captionfont}{\footnotesize}
    \centering
    \vspace{-10pt}
    \includegraphics[width=0.39\textwidth]{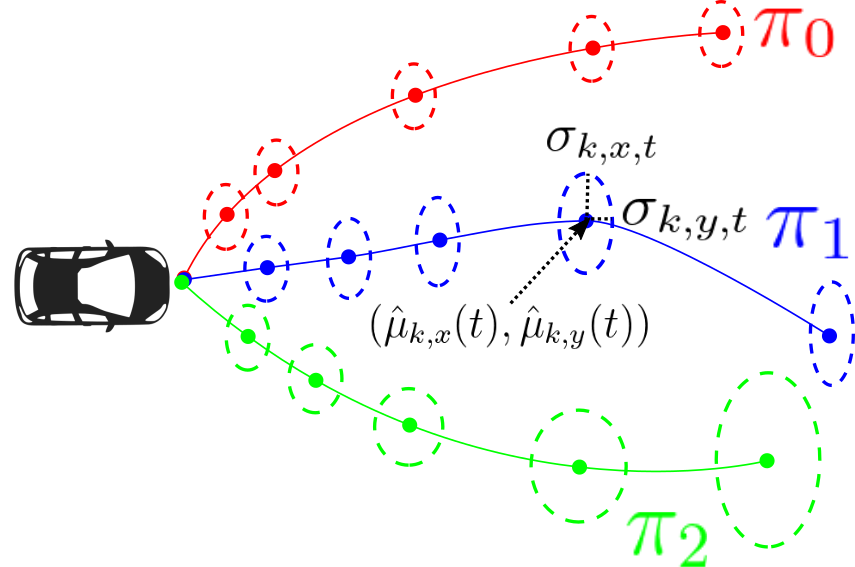}
    \caption{\textbf{Simplified representation of the trajectory prediction model ($K=3, T=5$).} Each sampled Gaussian component is represented by an ellipse of center $\big(\hat{\mu}_{k,x}(t), \hat{\mu}_{k, y}(t) \big)$ and shape $(\sigma_{k, x, t}, \sigma_{k, y, t})$.}
    \label{fig:trajs}
\end{wrapfigure}

Finally, for each sampled point in the future at time $t \in [0, T]$, the probability density function for the point position, 
$p(x, t)$
is expressed as follows, for each vehicle 
($p(\cdot|c)$ for the ego vehicle, for each command): $p(x, t) = \sum_{k=1}^{K} \pi_k \mathcal{N}\big(\hat{\mu}_{k, x}(t), \sigma_{k, x, t} \big)$
where $\hat{\mu}_{k, x}(t) = \sum_{d=0}^{d=3} a_{k, d, x} t^{4 - d}$.
Expressions for $p(y, t)$ are analogous.

The outputs of the network are: the vectors $\boldsymbol{a}_{k, x}$ (resp. $\boldsymbol{a}_{k, y}$), of dimension $d$, the 
variance $\sigma_{k, x, t}$ (resp. $\sigma_{k, y, t}$) and global trajectory coefficients $\pi_k$ for each vehicle, and for each command $c$ in the case of the ego vehicle. 
In a nutshell, this representation can be interpreted as predicting $K$ trajectories, each associated with a confidence $\pi_k$, with sampled points following a Gaussian distribution centered on $\big(\hat{\mu}_{k,x}(t), \hat{\mu}_{k, y}(t) \big)$ (generated and constrained by polynomials) and with 
variance $(\sigma_{k, x, t}, \sigma_{k, y, t})$. 
We illustrate the idea through a simplified example in Figure~\ref{fig:trajs}.

\paragraph{Auxiliary semantic segmentation output}

The segmentation module predicts over 7 typical classes for visual perception-based driving (\textit{Void, Vehicle, Pedestrian, Traffic Sign/Signal, Lane Marking, Road, Sidewalk}).
Our objective here is to make sure that the features of the image encoder comprise additional information about the road position and availability, the applicability of the traffic rules (traffic sign/signal), the vulnerable road users (pedestrians, cyclists, etc.) position, etc.
This information is useful for trajectory prediction and can potentially contribute to  the explainability of the model.

\paragraph{Loss}

To train the network, the main objective of predicting the trajectories distribution is achieved by minimizing negative log-likelihood (NLL) over all sampled points of the ground truth ego and neighbor vehicles trajectories (see Equation~\ref{eq:loss-nll}, where $p_{\mathtt{ego}}$ is the distribution for the ego vehicle and $p_{n}$ the distribution for the $n$-th neighbor). To improve the results for the lateral component of the trajectories, which is harder to predict, we add a weight $\alpha$ to the loss related to the $y$ coordinate ($\alpha=3$ in our tests). In the end, the loss $L_{\mathtt{NLL}}$ is expressed by Equation \ref{eq:loss-nll}, where $(\overline{x}(t), \overline{y}(t))$ represents the ground truth position of the ego vehicle at time $t$, respectively $(\overline{x}_n(t), \overline{y}_n(t))$ for the $n$-th neighbor, and $c$ is the current navigation command for the ego vehicle:

\begin{equation}
    \small
    L_{\mathtt{NLL}}(c) = -\sum_{t=1}^{T} \left( \log \big( p_{\mathtt{ego}}(\overline{x}(t)|c)\big) + \alpha \log\big( p_{\mathtt{ego}}(\overline{y}(t)|c)\big) + \sum_{n=1}^{N} \Big[ \log \big( p_{n}(\overline{x}_{n}(t))\big) + \alpha \log \big( p_{n}(\overline{y}_{n}(t))\big) \Big] \right).
\label{eq:loss-nll}
\end{equation}

For the auxiliary semantic segmentation module, we use the cross-entropy loss $L_{\mathtt{seg}}$.
Since the considered datasets do not offer both semantic segmentation and object tracking annotations, we train the network by mixing samples containing either one of the objectives, and backpropagate using either $L_{\mathtt{NLL}}$ or $L_{\mathtt{seg}}$. For training, we used Radam optimizer \citep{RAdam}, which is an improvement of Adam, with learning rate $10^{-5}$. The mini-batch size was set to 8 and each mini-batch was split in two: 4 semantic segmentation task inputs and 4 trajectory task inputs.

\section{Experiments on Offline Data}

\subsection{Metrics}

We use two main metrics: minMSD (minimum Mean Squared Deviation) as in \citep{precog, r2p2, desire} and minADE (minimum Average Displacement Error): 
$\text{minMSD} = \frac{1}{TN}\sum_{n=1}^{N} \min_{k \in K} \sum_{t=1}^{T} ||\mu_{n,k}(t)-\mu_n^*(t)||^2$
and $\text{minADE} = \frac{1}{TN}\sum_{n=1}^{N} \min_{k \in K} \sum_{t=1}^{T} ||\mu_{n,k}(t)-\mu_n^*(t)||$.

These metrics consider only the best predicted trajectory for the selected metric regardless of its confidence score. The purpose is to avoid penalizing valid possible future trajectories for each agent when it does not correspond to the actual recorded path. For example, penalizing a prediction that turns right with high confidence and goes straight with smaller but still high confidence when the vehicle went straight during the recording is not appropriate. minMSD will emphasize high errors while minADE is neutral regarding the error magnitude.

\subsection{Comparison to State of the Art}

\begin{wraptable}[10]{r}{0.45\textwidth}
    \renewcommand{\captionfont}{\footnotesize}
    \centering
    \scriptsize
    \vspace{-10pt}
    \begin{tabular}{p{1.9cm}|p{0.18cm}*{5}{p{0.15cm}}}
         \toprule
         & \multicolumn{6}{c}{minMSD ($m^2$)} \\
         Number of agents & 1 & 2 & 3 & 4 & 5 & 6+  \\
         \midrule
         DESIRE-plan~\citep{desire} & 2.26 & 6.64 & 6.18 & 9.20 & 8.52 & -\\
         ESP~\citep{precog} & 1.86 & 2.37 & 2.81 & 3.20 & 4.36 & -\\
         PRECOG~\citep{precog} & \textbf{0.149} & 2.32 & 2.65 & 3.16 & 4.25 & - \\
         PLOP & 1.89 & \textbf{1.97} & \textbf{2.39} & \textbf{2.74} & \textbf{2.84} & 2.53 \\
         \bottomrule
    \end{tabular}
    \caption{Comparison with published results of DESIRE-plan, ESP and PRECOG from \citep{precog} (results from their Table~II, with a fixed 5 agents training), over minMSD metric.}
    \label{tab:baselines}
\end{wraptable}

We conduct all experiments on two recent datasets: nuScenes ~\citep{nuscenes} and A2D2~\citep{aev2019}.
nuScenes is used as the trajectory prediction dataset, it consists of 850 scenes of 20s driven in Boston and Singapore. The 2Hz annotation of the scenes (tracked bounding boxes) is extended to 10Hz using interpolation.
We use the nuScenes train set as a train + validation set and the nuScenes validation set as a test set. We use only a small part of the Audi A2D2 dataset, namely image semantic segmentation data. We take the 41,280 annotated frames and reduce the available classes to the 7 classes we consider. At training time, the nuScenes examples are used only with trajectory losses and the examples from Audi with semantic segmentation loss. We evaluate our results only on the nuScenes dataset since the semantic segmentation task is auxiliary. The goal of semantic segmentation here is not accuracy, but rather to implicitly inject awareness about the scene layout into the network.

For comparison, we use the results published in PRECOG \citep{precog}, with both ESP and PRECOG itself, and also report some of their baselines for comparison. We distinguish the evaluation of the trajectory prediction regarding the number of agents in the current scene, from one agent (ego vehicle only) up to 10 agents (9 neighbors).
There is information on the goal of the ego vehicle but not for the others. %
Note that in our setup, the goal is given as a high-level navigation command, whereas PRECOG gives the goal as the target position 4s ahead, which puts PLOP at a slight disadvantage here.
The comparison is fairer for neighbor trajectories, and the performance is relevant as they are by definition open loop.

Results are presented in Table~\ref{tab:baselines}. The comparison is made using minMSD, as reported in PRECOG \citep{precog}.
We note that ESP and PRECOG are not fully flexible concerning the number of agents considered, we choose to compare to the published results trained for a maximum of 5 neighbors. To reach maximum performance using ESP and PRECOG, it is required to train specifically for each situation: 1 agent, 2 agents, etc. For a fair comparison, all results presented in Table \ref{tab:baselines} are computed for $K=12$.
Considering ego trajectory prediction only (1 agent), PRECOG outperforms our architecture
significantly. This is expected,
since the navigation goal is given as an exact target position, 
compared with our high-level navigation command. However, PLOP performs as well as ESP, even if ESP has access to the same goal as PRECOG.
For situations with up to 4 agents, PLOP slightly outperforms others
Finally, for 5 agents or more, our method outperforms by a large margin all other approaches.
This shows that, unlike compared methods, our model is 
more robust to the varying number of neighbors: note that with 3 or more agents, the minMSD metric, regarding the neighbors vehicles only, varies 
little. This result might be explained by our interaction encoding which is robust to the variations of $N$ using only multiple BEV projections and our non-iterative single step trajectory generation. 
We believe this is a valuable results as running one model per situation in a real car poses major technical challenges.
The metric improves while going from 5 to 6 or more agents, such crowded situations often involve slow or stopped vehicles which are easier to predict.

\subsection{Finer analysis on metrics and architecture}

\begin{wraptable}[6]{r}{0.4\textwidth}
    \renewcommand{\captionfont}{\footnotesize}
    \centering
    \vspace{-10pt}
    \scriptsize
    \begin{tabular}{p{1cm}*{6}{p{0.25cm}}}
        \toprule
        Number of agents & 1 & 2 & 3 & 4 & 5 & 6+ \\
        \midrule
        minADE & 0.85 & 0.84 & 0.90 & 0.94 & 0.90 & 0.76 \\
        \bottomrule
    \end{tabular}
    \caption{Comparison of minADE ($m$) metric according to the number of agents.}
    \label{tab:full_minade}
\end{wraptable}

Table~\ref{tab:full_minade} reports minADE. We notice that minADE values for the different situations are 
overall close to each other. We note that, due to the square factor, small variations in the minADE metric can induce higher variations in the minMSD metric. We also note that low minADE value along high minMSD values tends to represent a distribution with few very incorrect trajectories and a great number of correct trajectories, whereas high minADE value along low minMSD value tends to represent a distribution constituted of a lot of moderately correct trajectories.

\begin{wraptable}[9]{r}{0.5\textwidth}
\renewcommand{\captionfont}{\footnotesize}
\centering
\scriptsize
\vspace{-10pt}
\begin{tabulary}{0.5\textwidth}{J|CC|CC}
\toprule
 & \multicolumn{2}{c}{Ego vehicle} & \multicolumn{2}{c}{Neighbor vehicles} \\
 \midrule
 & minMSD & minADE & minMSD & minADE \\
\midrule
$K=12$ & \textbf{1.65} & \textbf{0.79} & \textbf{2.82} & \textbf{0.88}\\
$K=6$ & 2.28 & 0.90 & 3.77 & 1.05\\
$K=3$ & 2.55 & 0.93 & 6.81 & 1.44 \\
$K=1$ & 4.13 & 1.26 & 9.91 & 1.83\\
\bottomrule
\end{tabulary}
\caption{Influence of number of components for the Gaussian mixture $K$ over the metrics.}
\label{tab:traj_number}
\end{wraptable}

The number of predicted trajectories $K$ is fixed in the network architecture and we need to estimate the best value for this parameter. The presented results so far used $K=12$ for a fair comparison with other methods, however in other works up to one hundred or more trajectories per agent are predicted~\citep{desire}.
Observed results tend to show that increasing $K$ improves the metrics results. %
However, to keep a reasonable number of parameters in the output layer considering our trajectory generation we keep $K$ under 12.  The results of this hyperparameter study are presented in Table~\ref{tab:traj_number}. It confirms that adding more trajectory components improves the performance, although we can see that the contribution diminishes for $K \geq 6$.

\section{Closed Loop on Real Vehicle}

\subsection{Setup}

\begin{wraptable}[8]{r}{0.4\textwidth}
    \renewcommand{\captionfont}{\footnotesize}
    \centering
    \scriptsize
    \vspace{-10pt}
    \begin{tabulary}{0.4\textwidth}{J|CCCCC}
    \toprule
        Scenario & $\overline{S}$ & $\overline{N}$ & $\Sigma_N$ & $\Sigma_D$ & $\Sigma_T$ \\ %
    \midrule
        Urban & 19 & 5.6 & 13619 & 102 & 5h20 \\ %
        Track  & 17 & 0.5 & 369 & 43 & 2h26 \\ %
    \bottomrule
    \end{tabulary}
    \caption{Datasets characteristics: mean speed $\overline{S}$ (kph), mean number of neighbors per frame $\overline{N}$, total tracked neighbhors $\Sigma_N$, total distance $\Sigma_D$ (km) and time $\Sigma_T$}
    \label{tab:dataset}
\end{wraptable}

To alleviate the limitations of IL, it is necessary to use multiple data augmentation techniques. The goal is to produce scenarios outside of the training data, such as, lateral shift, lane departure, breaking anticipation or recovery. The methods are out of the scope of this work, but are detailed extensively in the literature \cite{nvidia, chauffeutnet, ConditionalVT, cil}. We also add noise in the vehicles positions (ego and neighbors) to improve our method robustness following a similar approach to \citep{ConditionalVT, Toromanoff_2020_CVPR}. We use a test vehicle equipped with high precision GPS, Lidar and surround-view cameras (replacing the frontal camera from nuScenes). GPS tracks are used as ground truth for the ego vehicle trajectories. For neighbors, we use an off-the-shelf vehicle detector \cite{chabot2017deep} and tracker, running at 25Hz, which is not manually annotated and thus contains noise. The dataset contains two types of scenes: a busy urban driving scenario and a test track covering multiple urban situations (traffic light, roundabout, yield, stop, etc) with only one other vehicle. The characteristics of this internal dataset are summarized in Table~\ref{tab:dataset} and detailed in the supplementary material. The prediction horizon is reduced to 2s ($T=20$) to correspond better to the control algorithm.

\subsection{Evaluation}

A closed loop setup is critical to evaluate trajectory prediction, especially for a behavioral cloning approach (see \citep{offlineimitation}). For example, on an offline benchmark, a trajectory prediction that entirely ignores traffic lights would be penalized only in the few frames where the vehicle braking should be anticipated. There are two possible ways to evaluate trajectory prediction for vehicle control: simulation, and running tests on the actual target vehicle. \citet{amini2020learning} present a thorough example: they use a data-driven simulator to train and evaluate driving policies offline, then for online behavior near-crash recoveries, trajectory variance and distance to mean trajectories are reported over a 3km test track. Our online experiments are conducted using a in-house simulator where we approximate real driving using a bicycle model for ego displacement and sensor transformations \cite{nvidia, lat_fisheye}. We have also performed preliminary tests in a real vehicle with the ego trajectory prediction on the urban test track. More information about our online tests is presented in the supplementary material.

\begin{wraptable}[12]{r}{0.4\textwidth}
    \renewcommand{\captionfont}{\footnotesize}
    \centering
    \scriptsize
    \begin{tabular}{p{1.5cm}|*{3}{m{0.2cm}}|*{3}{m{0.2cm}}}
    \toprule
         &  \multicolumn{3}{c}{Urban} & \multicolumn{3}{c}{Track}\\
         Failures & \textit{lat} & \textit{high} & \textit{low} & \textit{lat} & \textit{high} & \textit{low} \\
         \midrule
         PLOP & 0 & 40 & 1 & 4 & 7 & 2\\
         PLOP no seg. & 4 & 80 & 0 & 7 & 18 & 0 \\
         Const. vel. & 185 & 35 & 10 & 381 & 14 & 4\\
         MLP & 431 & 44 & 15 & 302 & 53 & 12\\
    \bottomrule
    \end{tabular}
    \caption{Quantitative comparison of failure cases, on the test set: PLOP, PLOP without semantic segmentation auxiliary loss (no seg.) vs two baselines, one assuming constant velocity (Const. vel.) and the other with a multi-layer Perceptron (MLP) }
    \label{tab:baseline_comparison}
\end{wraptable}

To evaluate performance in the simulator, we rely on 3 metrics: lateral, high speed and low speed errors count. Lateral errors (\textit{lat}) occur when the simulated vehicle deviates from the expert trajectory from more than 1m. High and low speed errors (\textit{high} and \textit{low}) occur when the simulated vehicle is too fast (catching up to a vehicle 15\% faster than the real vehicle up to 0.6s in the future) or too slow (simulated speed is 20kph or lower under the expert one). As expected, offline metrics such as minADE are not discriminating enough for the online behavior. Removing raw sensor data results roughly in a 10\% drop in minADE on nuScenes, and training a multi-layer perceptron (MLP) using only past trajectory data gives a final ADE of 0.4m on the test data. However these approaches are absurd online: they cannot access mandatory information such as traffic lights, road intersections etc. Similarly, the impact of semantic segmentation auxiliary loss can only be seen online. Quantitative results in the simulator over the test recordings, 8km of busy urban scenario and 5kms of test track, can be seen in Table~\ref{tab:baseline_comparison}, and errors locations can be visualized in Figure~\ref{fig:errors_pos}. More details are available in the supplementary materials.

\begin{figure}
    \centering
    \begin{subfigure}{0.24\textwidth}
    \includegraphics[height=4cm]{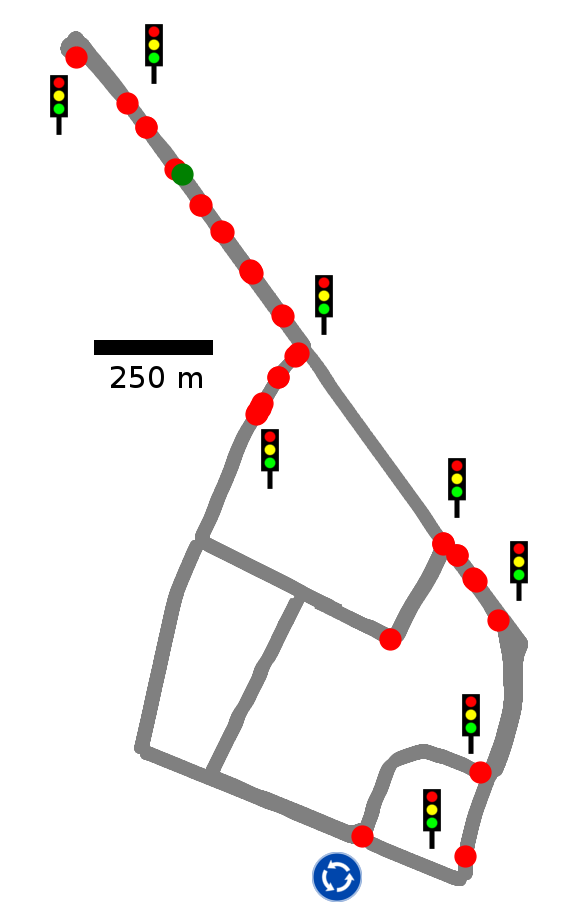}
    \caption{PLOP, urban}
    \end{subfigure}
    \begin{subfigure}{0.24\textwidth}
    \includegraphics[height=4cm]{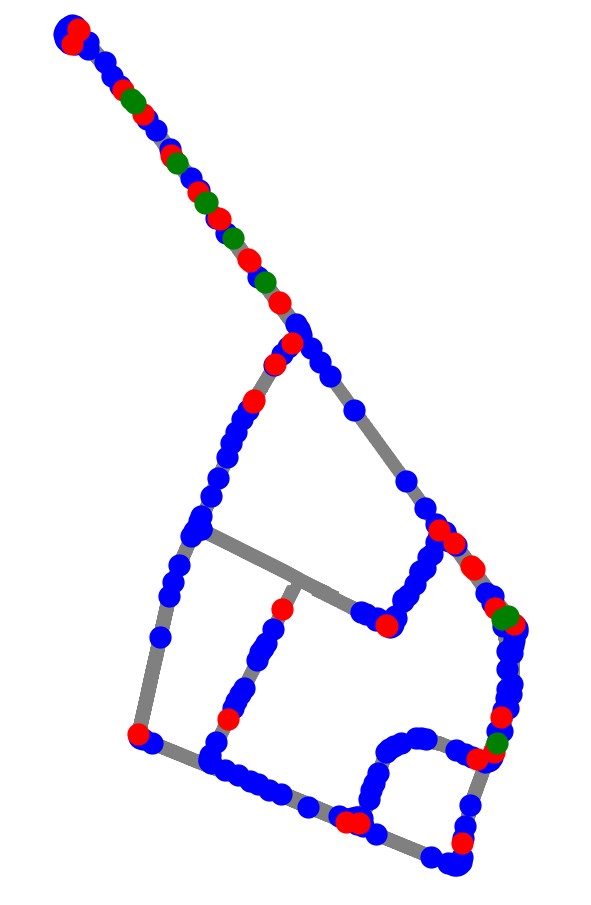}
    \caption{Const. vel., urban}
    \end{subfigure}
    \begin{subfigure}{0.24\textwidth}
    \includegraphics[height=4cm]{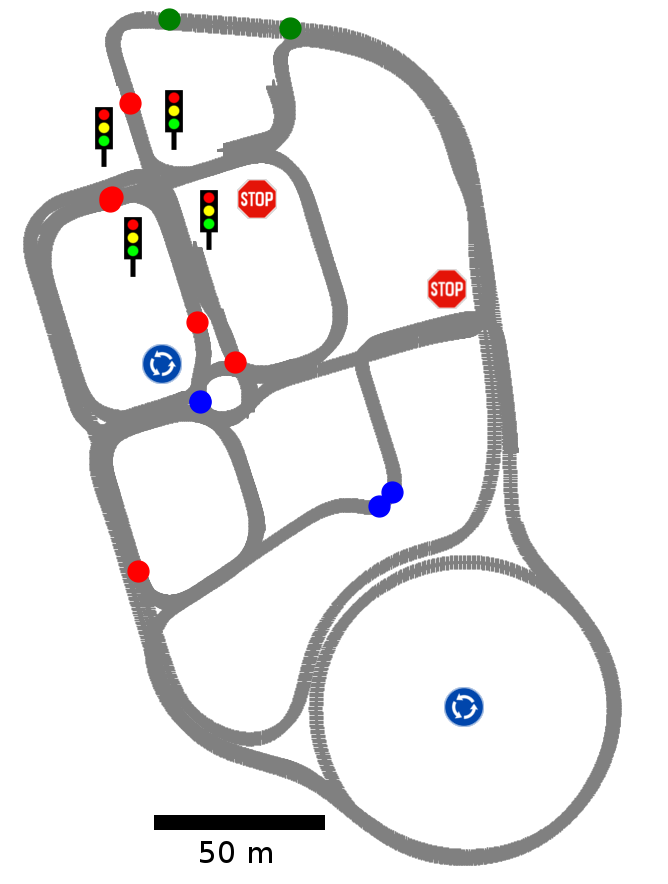}
    \caption{PLOP, track}
    \end{subfigure}
    \begin{subfigure}{0.24\textwidth}
    \includegraphics[height=4cm]{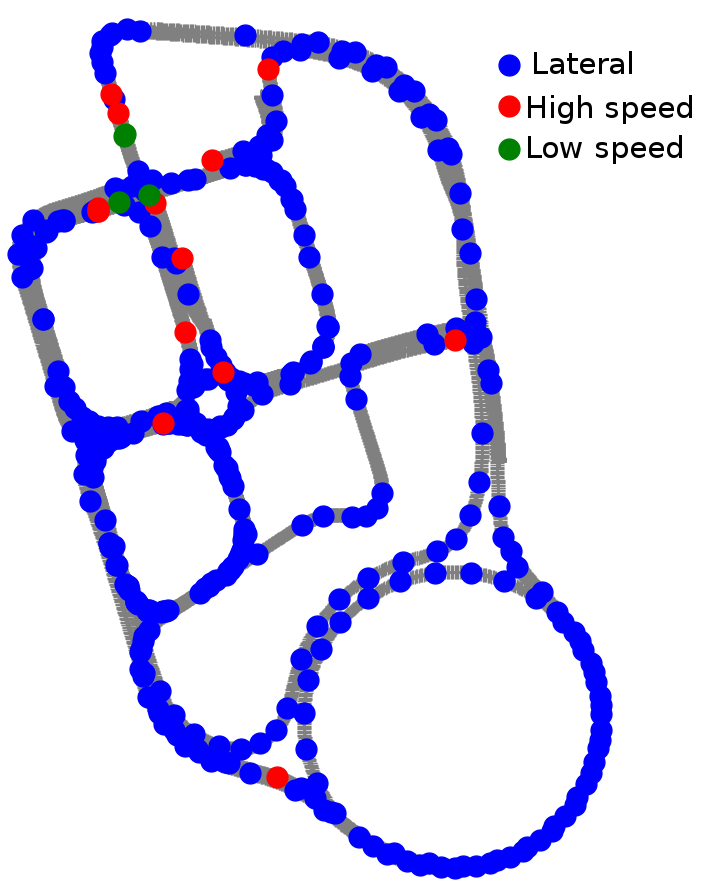}
    \caption{Const. vel., track}
    \end{subfigure}
    \caption{Closed loop error locations for urban and track test data, visualized for PLOP and constant velocity baseline. 
    We note that braking behind a vehicle can induce multiple high speed errors and stack multiple red dots on the same location. Points of interest (traffic lights, roundabout, stop signs) are highlighted on the map.}
    \label{fig:errors_pos}
\end{figure}

\subsection{Runtime and Optimization}

We implemented the proposed architecture using Tensorflow. We used an embedded device as computing platform (32Tops at best) for on-board computation and optimized our model using the TF-TRT library to use the full potential of the embedded GPU. We achieved 13-15 FPS online while returning all outputs (ego trajectories, neighbors trajectories and semantic segmentation) and were able to push the numbers to 22-25 FPS with only ego trajectories and neighbors trajectories outputs.

\section{CONCLUSIONS}

In this work, we demonstrate the interest of our multi-input multimodal approach PLOP for vehicle trajectory prediction in an urban environment. Our architecture leverages frontal camera and Lidar inputs, to produce multiple trajectories using reparameterized Mixture Density Networks, with an auxiliary semantic segmentation task. We show that we can improve open loop state-of-the-art performance in a multi-agent system, by evaluating the vehicle trajectories from the nuScenes dataset. We also provide a simulated closed loop evaluation, to go towards real vehicle online application.

In future work, 
we would be interested to include object detections as part of the architecture, to make it truly end-to-end and to use only raw sensor data. Including other types of road users, such as pedestrians or cyclists, could also make the system safer in a busy urban environment. %
Finally, it would be relevant to address the problem of generalization, and domain shift in particular due to weather conditions.

\section*{Supplementary material: active testing}

\subsection*{Test vehicle}
\label{sec:test-veh}

\begin{figure}
    \centering
    \includegraphics[width=1.0\textwidth]{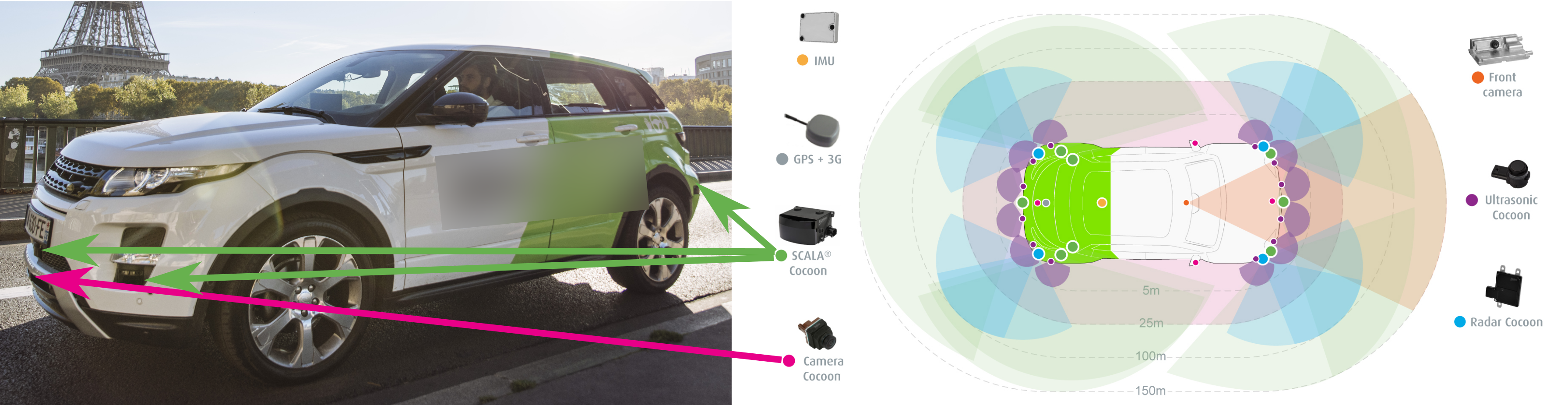}
    \caption{Test car setup, with the position of some of the Lidar sensors and the frontal camera (left), and sensor placement and field of view (right)}
    \label{fig:test_setup}
\end{figure}

The test vehicles are two Range Rover Evoque equipped with an industrial sensor suite (see Figure~\ref{fig:test_setup}). For our experiments, we rely on a frontal fisheye camera, situated just above the front license plate, and a 360 degrees array of Lidar sensors. The vehicle past positions are obtained from a high precision GPS and IMU.

During active testing, the predicted ego vehicle trajectory is converted to a steering and acceleration command using a control scheme designed to reach a point 1s in the future. This point is derived analytically from the trajectory polynom. The control algorithm is running on a dSpace microAutobox, and the actuation of pedals and steering wheel is done using a Paravan actuator.

\subsection*{Dataset}

\begin{wrapfigure}{r}{0.5\textwidth}
    \centering
    \includegraphics[width=0.5\textwidth]{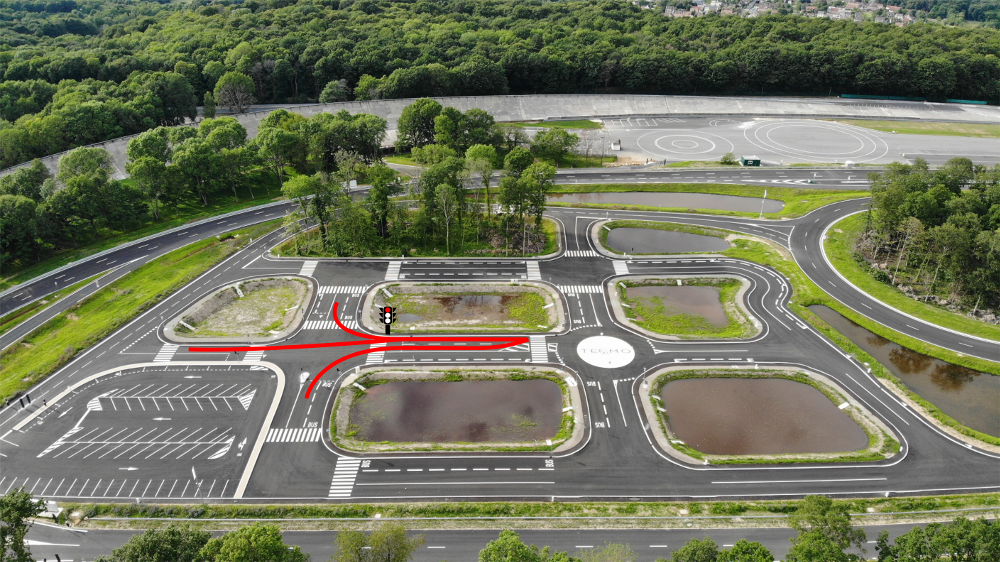}
    \caption{Aerial view of the test track with typical recording path overlayed}
    \label{fig:teqmo}
\end{wrapfigure}

We use the test vehicles (sensor calibration might be slightly different) to record driving data of two different scenarios over a one month period during the summer. First about 100km of driving were recorded in a complex real urban environment which can be challenging in many ways: erased markings, road users bad behavior, hidden signs, crowded streets, pedestrians, etc. The zone of driving was about 3 $km^2$ so each road was taken multiple times. Then 43 km of driving were recorded on a closed urban test track along with another vehicle (see Figure~\ref{fig:teqmo}). This is a less complex scenario in terms of randomness and other road users but denser in terms of signalisation and intersections. To avoid uselessly repetitive recordings over the test track, we used it to practice multiple scenarios such as yield, orange light, roundabout entry/exit, etc.

\subsection*{Ongoing active testing}

Before the pandemic, 2 days of online tests were performed on the test track with encouraging results. Main successes included lane changes, sharp turns, stop and go behind a vehicle and roundabout management. However, issues remained, in particular because of domain shift due to an important weather change between recording and testing.

Online tests in urban scenario have been delayed due to the pandemic situation but will be the focus once the situation improves. We will perform tests with increasing difficulty starting with straight lines and small curves, then going to intersection and traffic lights up to roundabouts \& crowded scenarios.

We present here a more detailed view of the network architecture, and additional insights on the evaluations of the main article, starting with the open loop scenario on nuScenes and finishing with the closed loop setup.

\section*{Detailed network architecture}

\begin{figure}[h]
    \centering
    \includegraphics[width=\textwidth]{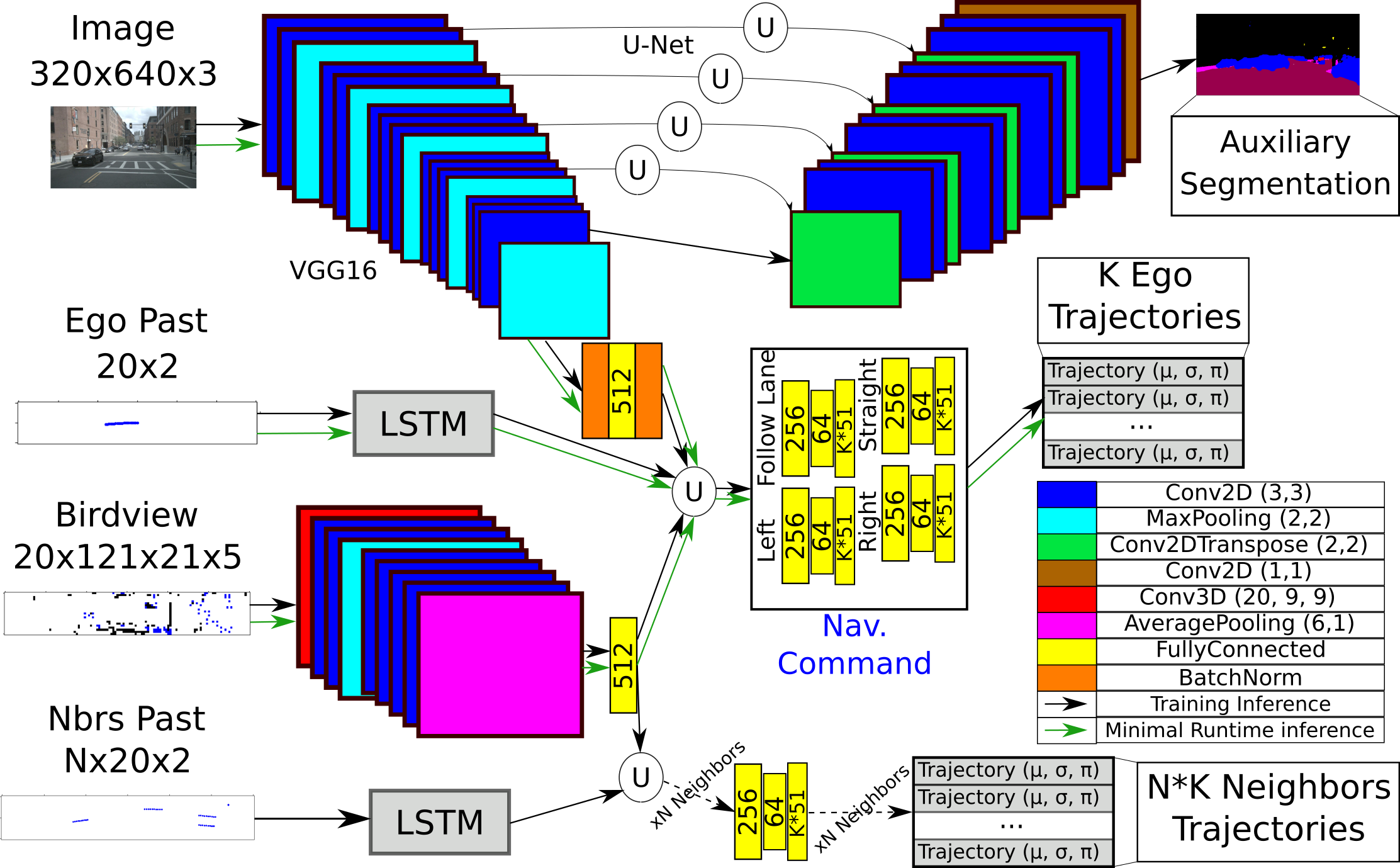}
    \caption{Detailed architecture of PLOP neural network}
    \label{fig:network-full}
\end{figure}

Figure~\ref{fig:network-full} shows the precise details of the network architecture of PLOP. In particular, the minimum path to predict the ego trajectory is highlighted: no semantic segmentation, no neighbor trajectories inputs. Note that there is no strict requirement on the image encoder backbone, so the VGG16 could be replaced by another encoder (Resnet, etc).

\newpage

\section*{Detailed results on nuScenes}

In this section, we present a finer analysis of the performance of PLOP on the nuScenes dataset. Other metrics are introduced to better study the behavior of the trajectory prediction. We also present the results of the ablation studies on the network structure. Finally, we present the performance of PLOP on a longer time horizon, in the context of a recent trajectory prediction benchmark of nuScenes.

\subsection*{Additional metrics}

In addition to minADE and minMSD, Final Displacement Error (FDE) (\ref{eq:minfde}) is interesting because it represents how good a method is at reaching a destination regardless of the intermediate path. minFDE is defined as follows:

\begin{align}
\begin{split}
    minFDE &= \frac{1}{N}\sum_{n=1}^{N} \min_{k \in K} ||\mu_{n,k}(T)-\mu_n^*(T)||.
\end{split}
\label{eq:minfde}
\end{align}

It can also be declined into other metrics such as a miss rate where we consider the final position missed if the FDE is greater than a distance threshold.

Planning multiple trajectories implies to select one of them, which cannot be evaluated using the minMSD and minADE metrics. For the neighbors, these metrics might be too lenient: a prediction where one trajectory fits perfectly the ground truth with an extremely low probability and all other trajectories poorly fit the ground truth might be considered as an inaccurate prediction while its minMSD and minADE are close to 0.
The selection problem is even more complex regarding ego trajectory. If we want to use it for control at some point, we might choose the trajectory with highest confidence given the current navigation command (no orientation issue), or we might try fusing multiple trajectories with high confidence into a safer one. To account for this, we also study the additional metrics confMSD, confADE and confFDE setting $k = \argmax_{k \in K} \pi_k$ in the corresponding equations to evaluate the relevance of the predicted confidence.
We also study \textit{weightFDE} (for weighted-FDE), defined as:

\begin{equation}
    weightFDE = \frac{1}{N}\sum_{n=1}^{N} \sum_{k=1}^{K} \pi_k ||\mu_{n,k}(T)-\mu_n^*(T)||.
\end{equation}

The goal of weightFDE is to check if high weight trajectory components also have a high error. Comparing it to confFDE highlights how alternative trajectories with non maximum weights compare to the most confident trajectory.

Finally, splitting the metrics regarding to the longitudinal axis $x$ or the lateral axis $y$ is also relevant. Depending on the situation an error on the $x$ axis might be negligible or on the contrary very significant compared to the same error on the $y$ axis.

\subsection*{Finer analysis on offline metrics}

\begin{figure}[t]
    \centering
    \begin{subfigure}{.4\textwidth}
      \centering
      \includegraphics[width=1.0\columnwidth]{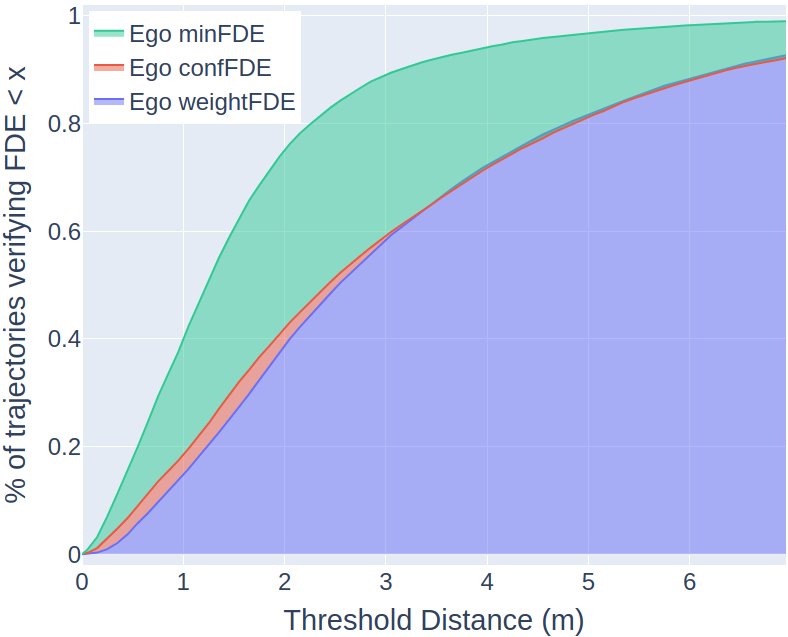}
      \caption{Ego Vehicle}
      \label{fig:ego_conf_min}
    \end{subfigure}%
    \begin{subfigure}{.4\textwidth}
      \centering
      \includegraphics[width=1.0\columnwidth]{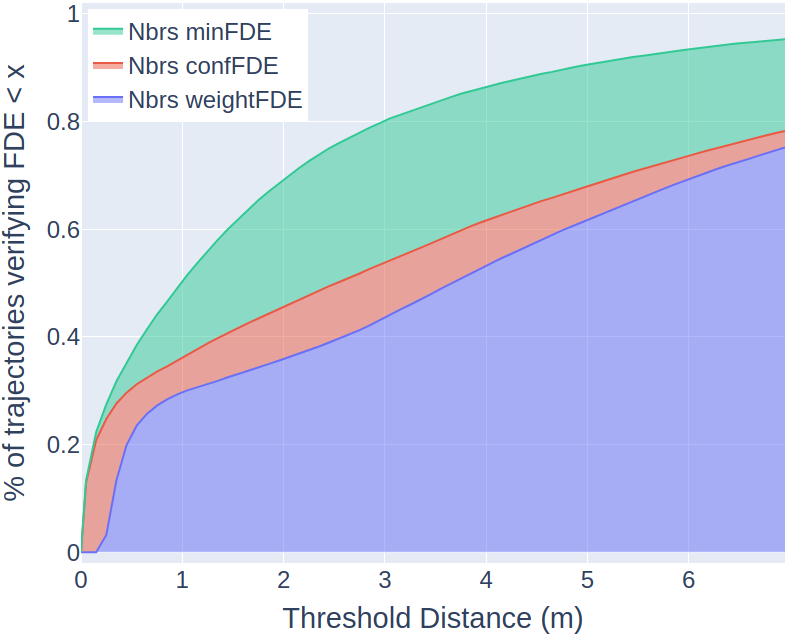}
      \caption{Neighbor Vehicles}
      \label{fig:nbrs_conf_min}
    \end{subfigure}

 \caption{Comparison between the $conf$, $weight$ and $min$ metrics. On both figures, the axis $x$ represents a threshold value in meters and on the $y$ axis value we report the accumulated percentage trajectories verifying $confFDE < x$, $weightFDE < x$ and $minFDE < x$ over the test set. Figure~\ref{fig:ego_conf_min}, resp. \ref{fig:nbrs_conf_min}, presents the results obtained for the ego vehicle only, resp. the neighbors only.}
 \label{fig:best_vs_min}
\end{figure}

\begin{figure}[t]
    \centering
    \begin{subfigure}{.4\textwidth}
      \centering
      \includegraphics[width=1.0\columnwidth]{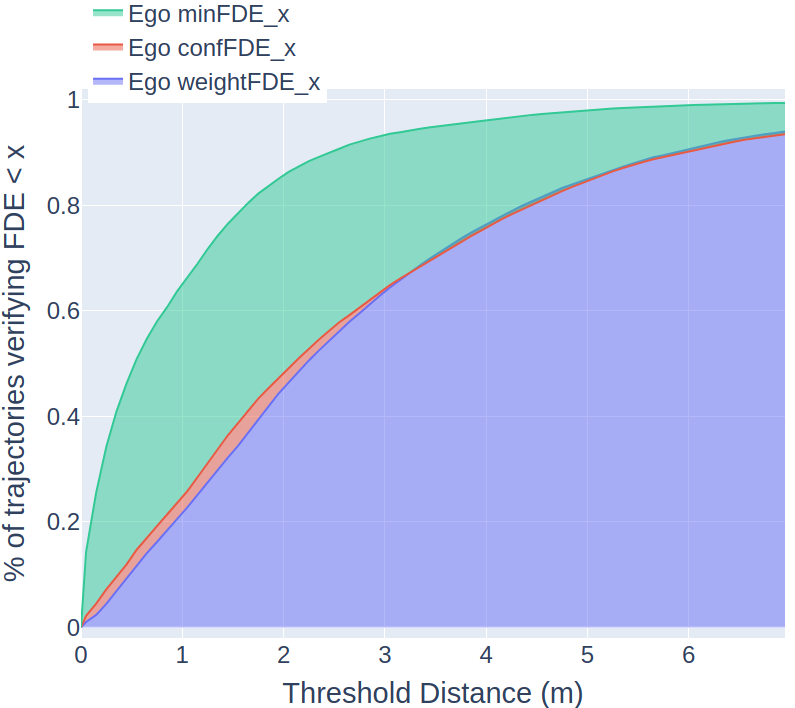}
      \caption{FDE over $x$ axis}
      \label{fig:ego_conf_min_x}
    \end{subfigure}%
    \begin{subfigure}{.4\textwidth}
      \centering
      \includegraphics[width=1.0\columnwidth]{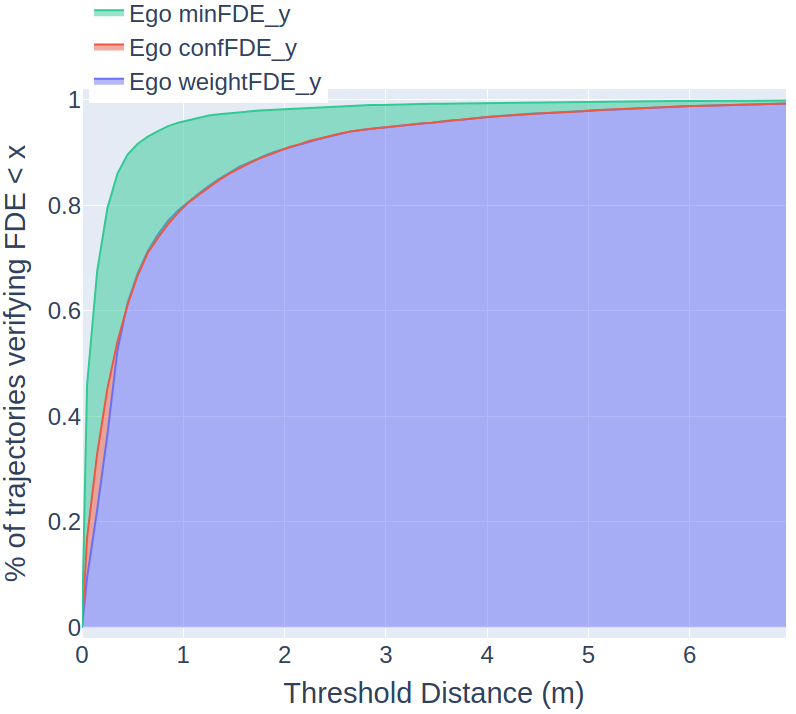}
      \caption{FDE over $y$ axis}
      \label{fig:ego_conf_min_y}
    \end{subfigure}

 \caption{Variations of FDE between the longitudinal and lateral axis. Plots are similar to \ref{fig:ego_conf_min}. The plot \ref{fig:ego_conf_min_x}, resp. \ref{fig:ego_conf_min_y}, compares the $minFDE$, $weightFDE$ and $confFDE$ metrics regarding the longitudinal axis, resp. the lateral axis, for the ego vehicle over the test set.}
 \label{fig:ego_conf_min_xy}
\end{figure}

We study the performance differences between the most confident trajectory, the confidence weighted sum of predictions and the closest to the ground truth trajectory using $conf$, $weight$ and $min$ metrics. The detailed comparison is presented in Figure \ref{fig:best_vs_min}: by definition $min$ yields better results that $conf$ and $weight$. 
For the neighbors, we note that the gap between the $min$ and $conf/weight$ curves is not closing up even when the FDE reaches 7m. This result exhibits the variety of behaviors in the predicted trajectories since similar behaviors have similar trajectories and vice versa. This result is not observed for the ego vehicle because it is conditioned by a navigation command.
 
Figure~\ref{fig:ego_conf_min_xy} illustrates the difference for the FDE metrics between $x$ and $y$ axis. We can see that FDE is significantly lower for $y$ than $x$. Even if ideally both errors should be as low as possible, we have to consider that the lateral error is the most critical for control and prediction.

\subsection*{Ablation studies on offline metrics}

To study the impact of our design choices, ablation studies are conducted on our network structure, both for inputs and auxiliary outputs. All networks are trained for a fixed number of 60 epochs. The full architecture is used as a benchmark architecture, and the following components are removed: Lidar input, camera input, semantic segmentation auxiliary loss, 
and finally multimodal trajectories (predicting only one single trajectory for each vehicle with $K=1$).

The results are presented in Table~\ref{tab:ablation} where each ablation row has to be compared to the \textit{Full} top row.
The ablation of the auxiliary semantic loss worsens the metrics for ego vehicle. It shows that teaching the network to represent such semantic in its features improves the prediction. This ablation also brought more instability during the training process.
Removing the camera does not yield results as poor as we could have expected for the ego vehicle trajectory planning, even if we cut out very useful information about the current scene like road markings, traffic light states, traffic signs, etc. An explanation could be the passivity of the testing method that does not penalize enough reactive behaviors (e.g. braking when the positions history is braking).
It would not be reasonable to cut out this kind of information in a real self driving car.
The two previous ablations also removed diversity in the observed data since they do not use the Audi semantic dataset for training. However, A2D2 also brings in some bias since it contains only right-hand drive data when nuScenes contains both: righ-hand and left-hand drive data.
We observe that removing the Lidar point cloud degrades all metrics, especially for the neighbors. This result was expected because we removed a lot of useful information about the free space of the static scene.
Using $K=1$ approach yields very poor results, also visible in the training loss. It was an anticipated outcome due to the ambiguity of human behavior.

\begin{figure}[t]
    \centering
    \begin{subfigure}{.4\textwidth}
      \centering
      \includegraphics[width=1.0\columnwidth]{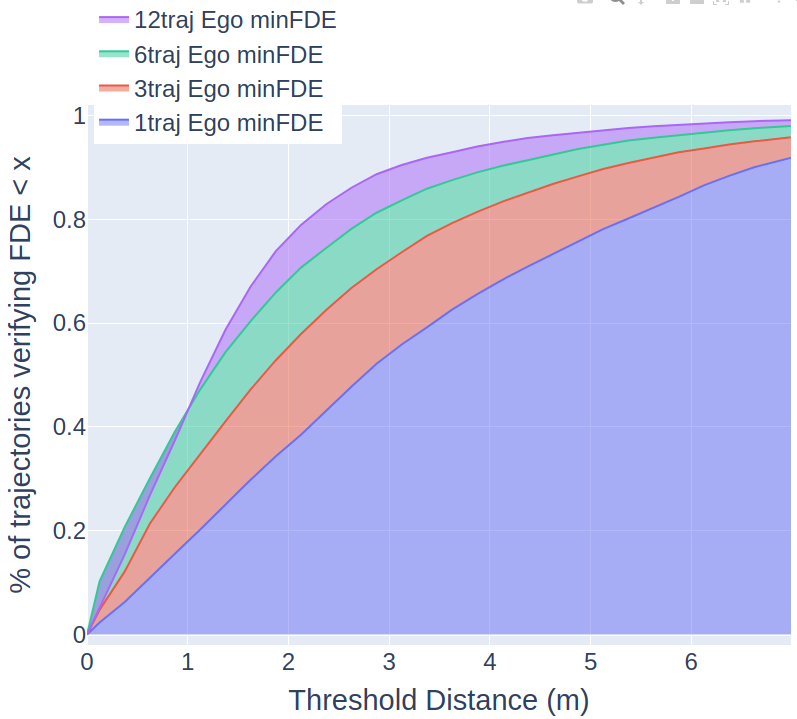}
      \caption{Ego Vehicle}
      \label{fig:ego_min_multitraj}
    \end{subfigure}%
    \begin{subfigure}{.4\textwidth}
      \centering
      \includegraphics[width=1.0\columnwidth]{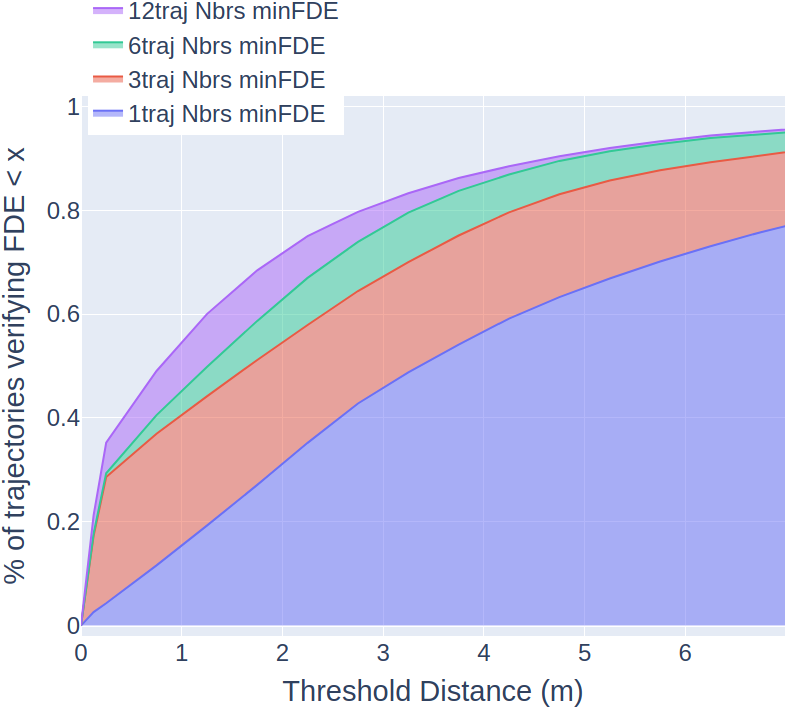}
      \caption{Neighbor Vehicles}
      \label{fig:nbrs_min_multitraj}
    \end{subfigure}

 \caption{Impact of the number of predicted trajectories $K$. Plots are similar to Figure~\ref{fig:best_vs_min}. The plot \ref{fig:ego_min_multitraj}, resp. \ref{fig:nbrs_min_multitraj}, presents the results obtained over the test set comparing the $minFDE$ performance using multiple values of $K$ for the ego vehicle only, resp. the neighbor vehicles only.}
 \label{fig:multitraj}
\end{figure}

Finally, the detailed impact of the number of components on minFDE can be visualized in Figure~\ref{fig:multitraj}. The results are consistent with those shown in Table~3 of the main article: adding new component has a positive yet decreasing impact on the performance. Interestingly, the difference is much steeper on the y axis than the x axis. This might be because there can be much more lateral variability on trajectories when at an intersection: in this kind of situations, more trajectories are required to model the possible behaviors.

\begin{table}
\centering
\begin{tabular}{c|cc|cc}
\toprule
 & \multicolumn{2}{c}{Ego vehicle} & \multicolumn{2}{c}{Neighbor vehicles} \\
 \midrule
 & minMSD & minADE & minMSD & minADE \\
\midrule
Full & \textbf{1.65} & 0.79 & \textbf{2.82} & \textbf{0.88}\\
No segmentation & 1.68 & \textbf{0.77} & 2.91 & 0.95\\
No camera & 1.72 & \textbf{0.77} & 2.95 & 0.90 \\
No Lidar & 1.80 & 0.81 & 3.02 & 0.93 \\
$K=1$ & 4.13 & 1.26 & 9.91 & 1.83\\
\bottomrule
\end{tabular}
\caption{Ablation study metrics for component removal.}
\label{tab:ablation}
\end{table}

\subsection*{Longer time horizon}

Recently, a benchmark on trajectory prediction has been introduced on the nuScenes dataset \footnote{\url{https://www.nuscenes.org/prediction?externalData=all&mapData=all&modalities=Any}}. This benchmark evaluates neighbors trajectory predictions over a 6s time horizon. Up to 25 future trajectories (\textit{modes}) can be proposed, and the following metrics are used: minimum ADE, FDE, miss rate at 2m (whether the predicted final point is in a 2m range of the ground truth). These metrics can be evaluated for the top $k$ modes, with $k=1, 5, 10$ (although only part of these results is publicly available, see Table~\ref{tab:nuscenes_6s}).

Our method is trained as described in the main paper, extending the prediction horizon to 6s (which means $T=60$), with the same number of maximum trajectories $K=12$. Trajectories are ranked according to their GMM weights $\pi_k$. Objects that are out of the input birdview range are predicted using the constant velocity baseline. We report our results in Table~\ref{tab:nuscenes_6s}. Note that this is far from ideal for our approach: a single polynomial of degree 4 is not enough to represent precisely represent such a long time horizon. Typically, the mean ADE between the ground truth positions and the polynomial fitting is around 0.6m, whereas for 4s it is around 0.2m.

\begin{table}[t]
    \centering
    \begin{tabular}{c|ccccccccc}
    \toprule
         & \multicolumn{3}{c}{minADE (m)} & \multicolumn{3}{c}{minFDE (m)} & \multicolumn{3}{c}{Miss rate (\%)} \\
         Top $k$ & $k = 1$ & $k=5$ & $k = 10$ & $k = 1$ & $k=5$ & $k = 10$ &  $k = 1$ & $k=5$ & $k = 10$ \\
         \midrule
         PLOP & 2.793 & 2.080 & 1.899 & \textbf{6.783} & 4.580 & 3.951 & 52.85 & 51.39 & \textbf{50.91} \\
         cxx & - & \textbf{1.630} & 1.288 & 8.865 & - & - & - & 68.95 & 60.47 \\
         MHA-JAM & - & 1.813 & \textbf{1.241} & 8.569 & - & - & - & 59.15 & \textbf{45.61}\\
         Physics oracle & - & 3.70 & 3.70 & 9.09 & - & - & - & 88.0 & 88.0 \\
         \bottomrule
    \end{tabular}
    \caption{Comparison over the nuScenes prediction benchmark: our approach (PLOP) compared to the top 2 rankings of the nuScenes prediction challenge (cxx and MHA-JAM) and a physical car model baseline.}
    \label{tab:nuscenes_6s}
\end{table}

For comparison, we present the results of the top 2 competitors at the time of submission (according to ADE and FDE): cxx \citep{luo2020probabilistic} and MHA-JAM \citep{messaoud2020multi}. Both methods rely on HD maps input. We also include the baseline \textit{Physics oracle} which is a trajectory prediction model that models the dynamics of the vehicle (more complex than the constant velocity baseline). Interestingly, PLOP is comparable to HD maps based methods. As expected, PLOP has worse results in $minADE$, which represents the general fitting of the trajectory. However, PLOP improves over both other methods in $minFDE$ and is competitive for miss rate, which shows that our approach can get better results on the final position estimation. We hypothesize that this is due to the fact that our prediction approach is single shot: the polynomial representation might be better at capturing the long term dynamic of the trajectory, at the cost of an approximation locally on the individual points. In contrast, sequential methods accumulate errors over time and do not have a global correction mechanism in the loss to compensate.

\newpage

\section*{Additional results on online simulation}

In this section, we give more details on the network input in the real vehicle setup. Then we add a slightly more detailed analysis of the closed loop simulated failure cases.

\subsection*{Details on image input}

\begin{figure}[h]
    \centering
    \subfloat[Original fisheye image]{\includegraphics[width=0.45\textwidth]{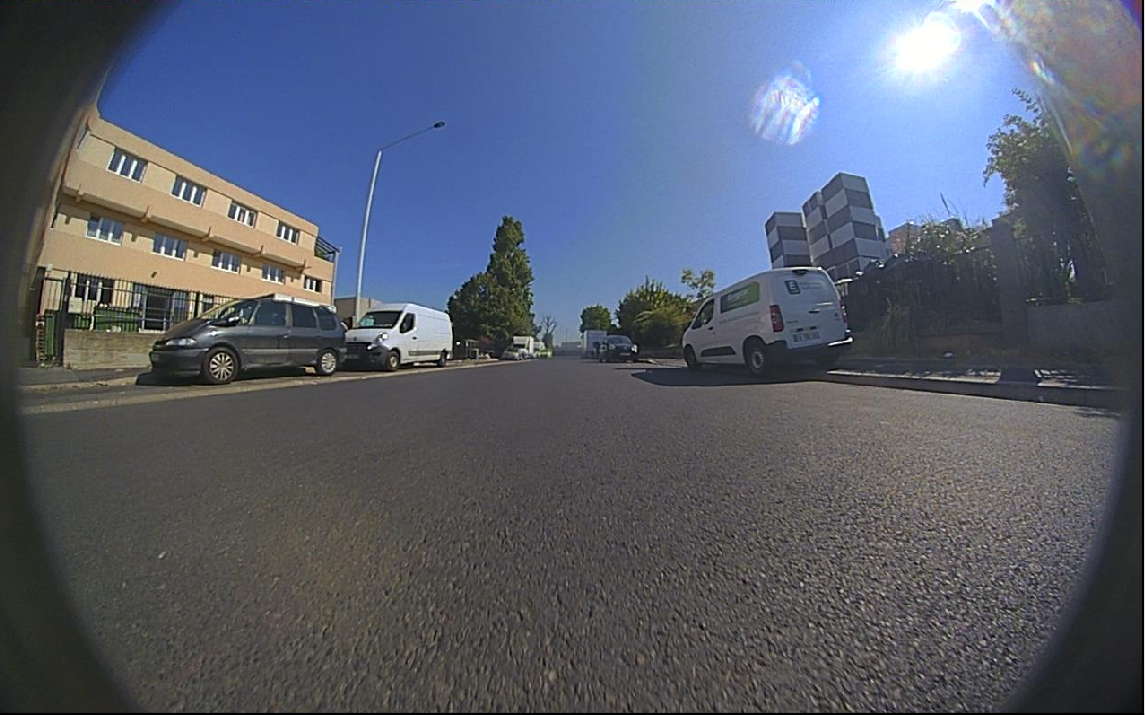}}
    \qquad
    \subfloat[Corresponding cylindrical reprojection]{\includegraphics[width=0.45\textwidth]{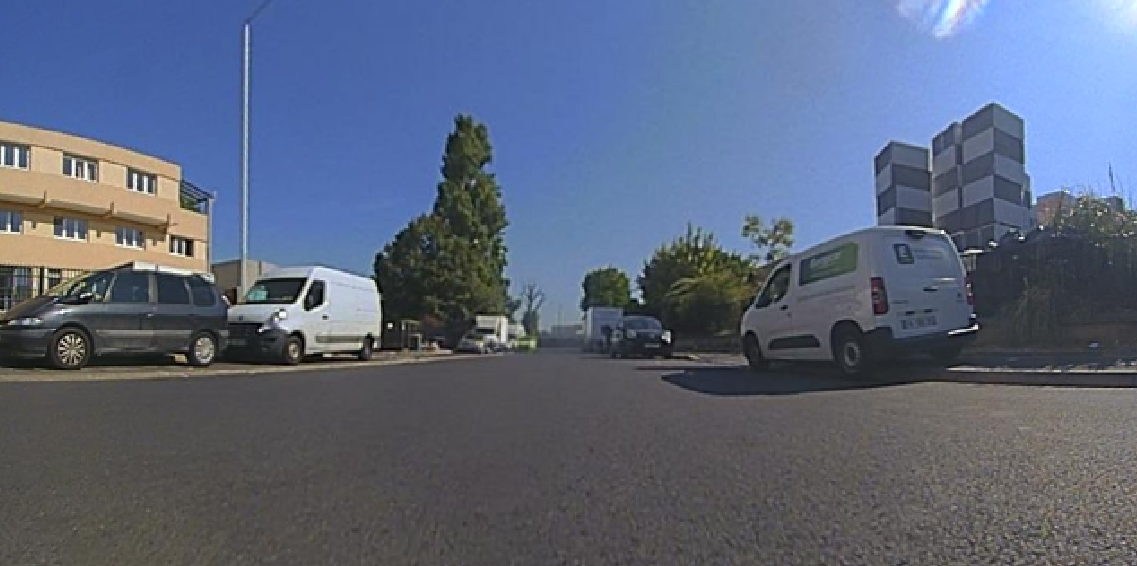}} \\
    \subfloat[Metric graduation on the original fisheye image]{\includegraphics[width=0.45\textwidth]{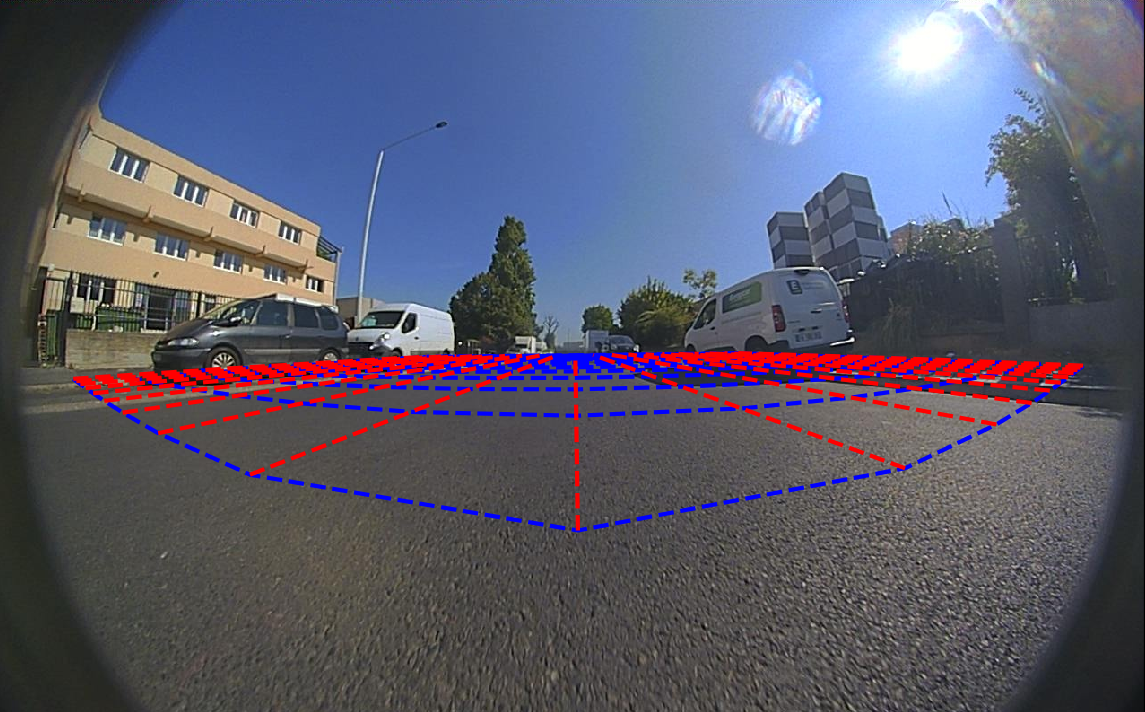}}
    \qquad
    \subfloat[Metric graduation on the cylindrical image]{\includegraphics[width=0.45\textwidth]{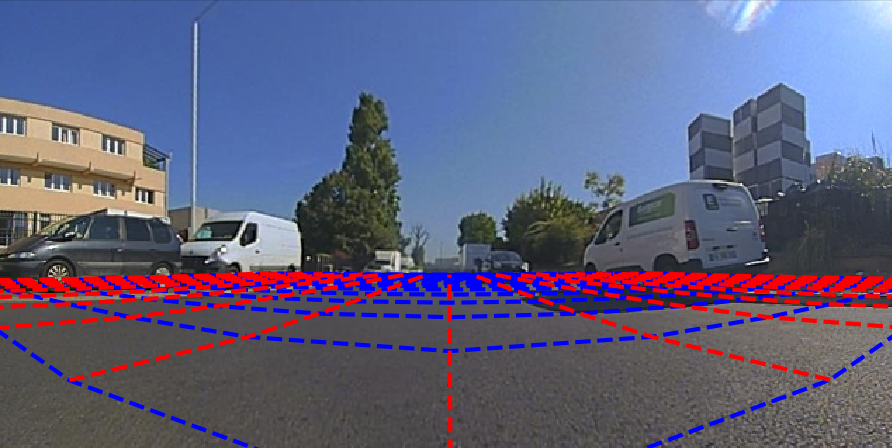}}
    \caption{Details on image input: original image vs cylindrical reprojection input, and visualization of detection range: $x \in [0.5; 20.5]m$ every 1m (blue dashed lines) and $y \in [-10; 10]m$ every 1m (red dashed lines)}
    \label{fig:fisheye}
\end{figure}

Figure~\ref{fig:fisheye} illustrates the image input data with more details. Here we present the raw, original fisheye images and compare them with the used cylindrical projection. The fisheye images correspond to 195 degrees of field of view, hence the strong effects of distortion. Note how the lines are straightened in the corresponding reprojection (150 degree field of view). This reprojection is done mainly to compensate for the distortion effects. To visualize the range of visibility in the image, we show lines corresponding to a ground plane projection spaced by 1m. In practice, the effective detection range is around 20-30m.

Using a fisheye camera is what makes it possible to generate deviations from the expert trajectories to correctly train the network with IL. This method is following the approached proposed by \citet{lat_fisheye} to generate the input images corresponding to deviations. For the Lidar data, the generation is much simpler as it is equivalent to a reference change in the point cloud. The drawback is that such cameras have a limited range, which advocates for the use of Lidar to compensate.

\subsection*{Closer analysis of closed loop failure cases}

\begin{figure}[h]
    \centering
    \begin{subfigure}{0.24\textwidth}
    \includegraphics[height=4cm]{images/urban_plop.png}
    \caption{PLOP, urban}
    \end{subfigure}
    \begin{subfigure}{0.24\textwidth}
    \includegraphics[height=4cm]{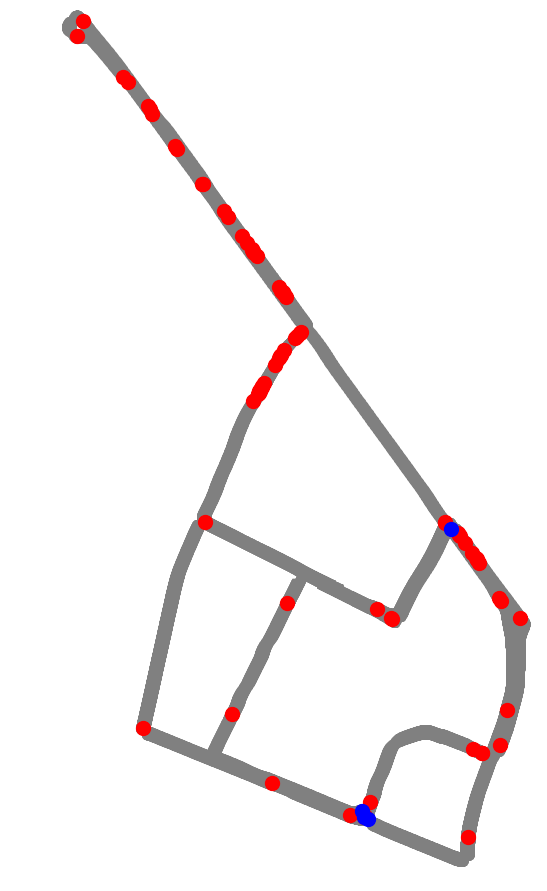}
    \caption{PLOP no seg., urban}
    \end{subfigure}
    \begin{subfigure}{0.24\textwidth}
    \includegraphics[height=4cm]{images/track_plop.png}
    \caption{PLOP, track}
    \end{subfigure}
    \begin{subfigure}{0.24\textwidth}
    \includegraphics[height=4cm]{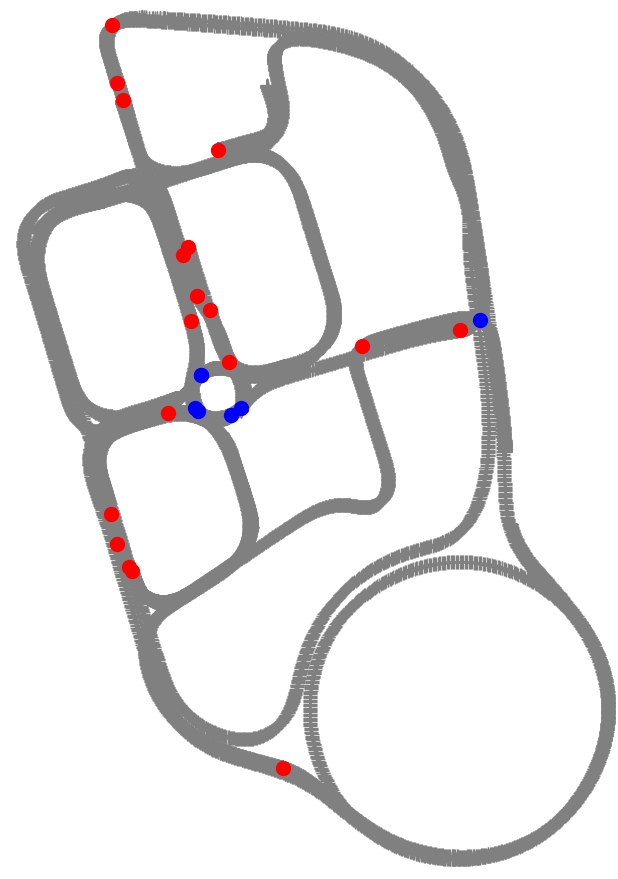}
    \caption{PLOP no seg., track}
    \end{subfigure}
    \caption{Closed loop error locations for urban and track test data, visualized for PLOP with or without semantic segmentation (no seg.) baseline. 
    }
    \label{fig:vs_no_seg}
\end{figure}

In Figure~\ref{fig:vs_no_seg} we present a detailed comparison between PLOP and PLOP trained without the semantic segmentation auxiliary loss. As seen in the main article, the online metrics are worse without the semantic segmentation. Interestingly, lateral errors seem to occur in difficult cases (hard turns, roundabout) without the semantic segmentation.

\begin{figure}[h]
    \centering
    \begin{subfigure}{0.3\textwidth}
    \includegraphics[height=5cm]{images/track_plop.png}
    \caption{PLOP}
    \end{subfigure}
    \begin{subfigure}{0.3\textwidth}
    \includegraphics[height=5cm]{images/track_cv.png}
    \caption{Constant velocity}
    \end{subfigure}
     \begin{subfigure}{0.3\textwidth}
    \includegraphics[height=5cm]{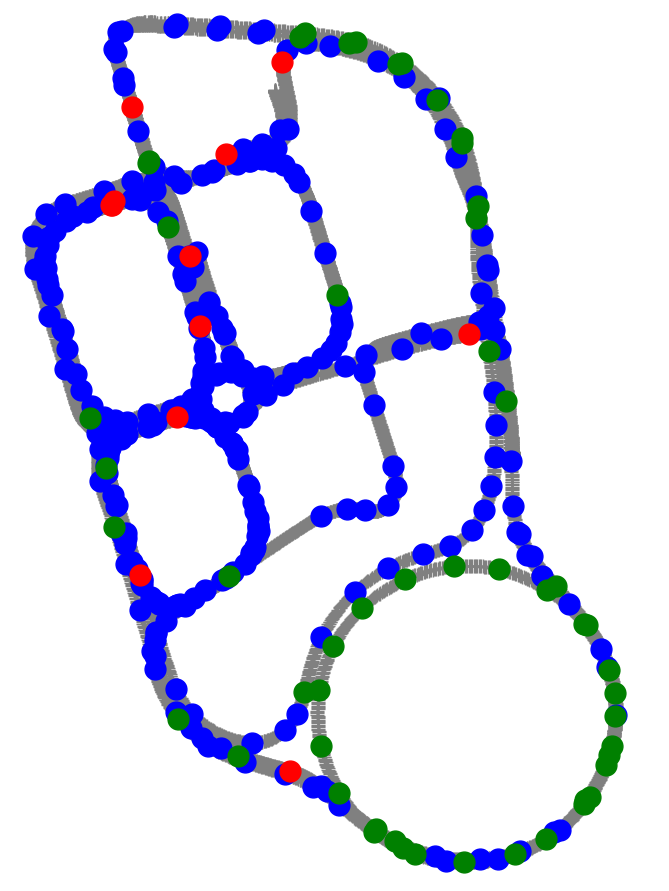}
    \caption{MLP}
    \end{subfigure}
    \caption{Detailed comparison of failure cases on track}
    \label{fig:detailed_track}
\end{figure}

Figure~\ref{fig:detailed_track} shows a more detailed view of the comparison of PLOP versus our baselines on the track setup. As expected, our baselines fail to stop at traffic lights and stop signs (high speed errors), because the input does not contain any information on the traffic light status. Interestingly, we can see that the failure cases for the lateral behavior are not the same between constant velocity and MLP prediction: in the constant curves (the large roundabout and the top-right curve), MLP actually manages to keep the vehicle in the lane, but fails on the speed. This is probably because the MLP is able to produce curved trajectories, but is not able to distinguish the driving speed between a sharp turn and a roundabout. 

\begin{figure}[h]
    \centering
    \subfloat[Missed traffic light (on track) due to sun glare.]{\includegraphics[width=0.45\textwidth]{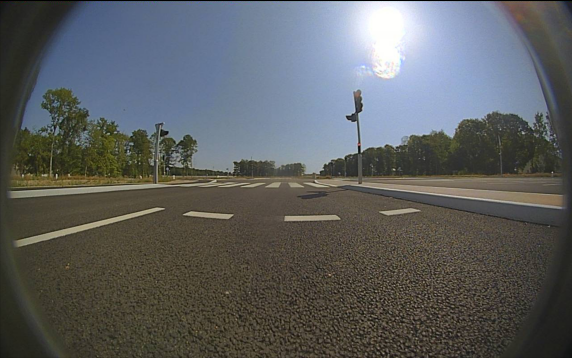}}
    \qquad
    \subfloat[Missed braking on a rare event of a truck stopped on roundabout.]{\includegraphics[width=0.45\textwidth]{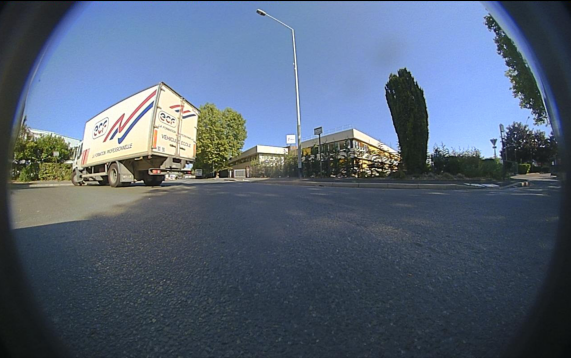}} \\
    \subfloat[Missed braking due to a sudden pedestrian crossing (far right).]{\includegraphics[width=0.45\textwidth]{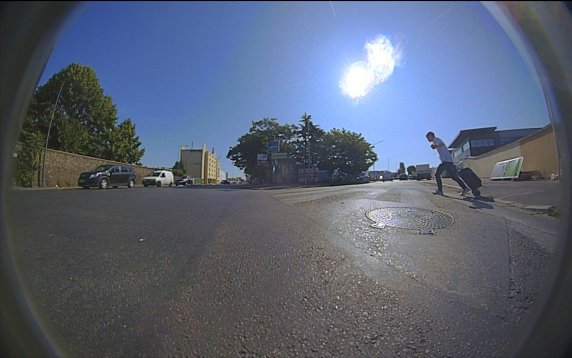}}
    \qquad
    \subfloat[Missed traffic light due to sun glare / partial obstruction by leaves.]{\includegraphics[width=0.45\textwidth]{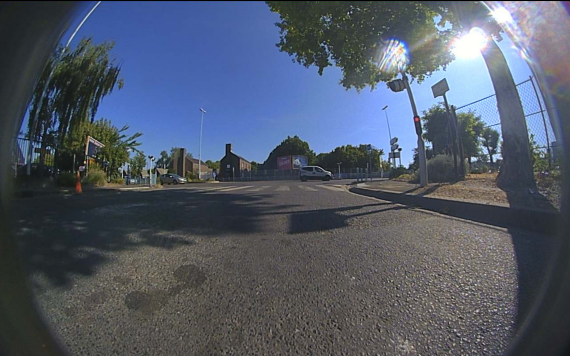}}
    \caption{Qualitative examples of failure cases because of high predicted speed (raw fisheye image)}
    \label{fig:failure_images}
\end{figure}

Figure~\ref{fig:failure_images} presents some qualitative examples of typical failure cases, where the speed predicted by PLOP is too high. Most braking errors come from difficult traffic lights, harder vehicles (trucks), or difficult use cases (non careful pedestrian). In future work, adding more cameras, typically from the side mirrors, might mitigate some of these use cases.

{
\small
\bibliography{plop,sup}
}

\end{document}